\pdfoutput=1

\documentclass[sigconf,nonacm]{acmart}

\usepackage{balance} %

\newif\ifpreprint
\usepackage{amssymb}
\usepackage{multirow}
\usepackage{nicefrac}
\usepackage{subcaption}
\usepackage{tikz}
\usepackage[capitalise,noabbrev]{cleveref}
\usepackage[colorinlistoftodos]{todonotes}
\usepackage{xspace}
\usepackage{algorithm}
\usepackage{algpseudocode}
\usepackage{listings}

\lstdefinestyle{promptstyle}{
  basicstyle=\ttfamily\footnotesize,
  frame=single,
  framerule=0.4pt,
  breaklines=true,
  breakindent=2em,
  breakatwhitespace=true,
  columns=flexible,
  keepspaces=true,
  xleftmargin=0.5em,
  xrightmargin=0.5em,
  aboveskip=1em,
  belowskip=1em,
  captionpos=b,
  abovecaptionskip=0.4em,
}

\crefname{lstlisting}{listing}{listings}
\Crefname{lstlisting}{Listing}{Listings}

\usetikzlibrary{positioning, arrows.meta, fit, backgrounds, calc, shapes.geometric, shapes.symbols, shapes.misc}

\usepackage{placeins} %

\newcommand{\vevt}[1]{\hat{V}_{#1}}

\newcommand{\rmstate}{u}

\newcommand{\themethodsparse}{RM-VLM}
\newcommand{\themethoddense}{RM-VLM+GVL}

\newtheorem{definition}{Definition}

\newtheorem{example}{Example}
\newtheorem{prop}{Proposition}
\crefname{prop}{Proposition}{Propositions}
\Crefname{prop}{Proposition}{Propositions}

\newcommand{\arbitraryset}{\mathcal{X}}

\newcommand{\dataset}{\mathcal{D}}
\newcommand{\labelleddataset}{\dataset_{\text{prop}}}
\newcommand{\trajectory}{\tau}

\newcommand{\rmtext}[1]{T_{#1}}
\newcommand{\taskset}{\mathcal{T}}

\usepackage{soul}

\newcommand{\reals}{\mathbb{R}}
\newcommand{\set}[1]{\{#1\}}
\newcommand{\simplex}[1]{\Delta(#1)}

\newcommand{\powerset}[1]{2^{#1}}

\newcommand{\mdp}{\mathcal{M}}

\newcommand{\initialdist}{\mu}
\newcommand{\states}{\mathcal{S}}
\newcommand{\statee}{s}
\newcommand{\actions}{\mathcal{A}}

\newcommand{\discountfactor}{\gamma}
\newcommand{\transitionprob}{p}

\newcommand{\propositions}{\mathcal{P}}
\newcommand{\llmpropositions}{\propositions}
\newcommand{\proposition}{p}
\newcommand{\labelingfunction}{\mathcal{L}}
\newcommand{\noisylabelingfunction}{\tilde{\labelingfunction}}

\newcommand{\propr}{\textcolor{red!75!black}{\mathsf{R}}}
\newcommand{\propg}{\textcolor{green!50!black}{\mathsf{G}}}
\newcommand{\propb}{\textcolor{blue!75!black}{\mathsf{B}}}
\newcommand{\propc}{\bigcirc}
\newcommand{\propt}{\triangle}

\newcommand{\rmstates}{U}
\newcommand{\initialrmstate}{\rmstate_0}
\newcommand{\acceptingrmstate}{\rmstate_A}
\newcommand{\rejectingrmstate}{\rmstate_R}
\newcommand{\rmstatetransitionfun}{\delta_u}
\newcommand{\rmrewardtransitionfun}{\delta_r}
\newcommand{\rewardmachine}{\mathcal{RM}}
\newcommand{\llmrewardmachine}{\rewardmachine}

\newcommand{\rmbelief}{\tilde{\rmstate}}
\newcommand{\labelobs}{L}
\newcommand{\noisylabelobs}{\tilde{L}}
\newcommand{\rmalpha}{\alpha}

\newcommand{\densereward}{r^{\text{dense}}}
\newcommand{\rmpotential}{v_{\rewardmachine}}
\newcommand{\rmdiscount}{\gamma_{\rewardmachine}}
\newcommand{\rmvalue}{\hat{V}_{\rewardmachine}}

\preprinttrue

\setcopyright{none}
\title{\themethodsparse{}: From Natural-Language Task Descriptions to Vision-Grounded Reward Machines}

\author{Roko Parac}
\affiliation{
  \institution{Imperial College London}
  \city{London}
  \country{United Kingdom}}
\email{roko.parac18@imperial.ac.uk}

\author{Christos Ziakas}
\affiliation{
  \institution{Imperial College London}
  \city{London}
  \country{United Kingdom}}
\email{c.ziakas24@imperial.ac.uk}

\author{Leo Ardon}
\affiliation{
  \institution{Imperial College London}
  \city{London}
  \country{United Kingdom}}
\email{leo.ardon19@imperial.ac.uk}

\author{Alessandra Russo}
\affiliation{
  \institution{Imperial College London}
  \city{London}
  \country{United Kingdom}}
\email{a.russo@imperial.ac.uk}

\begin{abstract}
Reward machines (RMs) have been proposed as a reward formulation in reinforcement learning (RL) to capture non-Markovian tasks where credit assignment depends on history.
By representing reward functions as automata, RMs expose the task's temporal structure and allow such rewards to be handled within a Markovian framework. 
However, they assume a labelling function, a proposition-grounding oracle for observations, which must be manually engineered for each environment, limiting their wider applicability. 
In this work, we address this limitation by proposing \themethodsparse{}, a novel approach in which a large language model (LLM) generates an RM from a natural-language task description, and a vision-language model (VLM) acts as the labelling function, grounding the RM's propositions over image-based observations during RM-guided policy learning. 
A VLM-based value function estimator is used to assess the agent’s progress towards each RM transition and derive a dense reward signal.
We evaluate \themethodsparse{} on three domains covering non-Markovian grid-world navigation, robotic manipulation, and 3D object collection. 
Our experimental results show that \themethodsparse{} consistently achieves higher task success rates than existing VLM-based reward methods across all tested domains.

\end{abstract}


\begin{document}
\pdfinfo{/Author (Roko Parac, Christos Ziakas, Leo Ardon, Alessandra Russo)}

\maketitle
\thispagestyle{plain}
\pagestyle{plain}

\section{Introduction}

Many reinforcement learning~\citep[RL;][]{sutton1998reinforcement} tasks require an agent to remember what it has done. 
An object may be worth collecting only after a prerequisite has been met; an action may be rewarding only at a particular stage of a task. 
In these settings, reward depends on the history of states and actions, rather than on the current observation alone. 
Such non-Markovian rewards make credit assignment depend on history, and remain difficult to specify and learn with standard RL methods.

\emph{Reward machines}~\citep[RMs;][]{ToroIcarteKVM18} offer a structured way to address this challenge. 
An RM represents a task as an automaton whose states compactly summarise task history and whose transitions capture progress through sub-goals. 
By augmenting the agent's state with the RM state, an RM makes the history relevant to reward explicit and can provide intermediate rewards as the task unfolds. 
Applying RMs, however, typically requires two forms of manual engineering: specifying the automaton and constructing a \emph{labelling function} that maps observations to the \emph{atomic propositions} governing its transitions. 
Labelling functions often rely on access to the environment's internal state, which may be unavailable in image-based or real-world settings.

Existing methods address parts of this problem, but retain significant dependencies. 
ARM-FM~\citep{castanyer2025arm} uses foundation models but implements the labelling function from the environment's source code. 
Ground-Compose-Reinforce~\citep[GCR;][]{li2025ground} learns a labelling function from trajectories, but requires manual labels. 
Methods such as ISA~\citep{furelos2021induction}, FORM~\citep{ardon2024form}, and Prob-IRM~\citep{parac2024learning} can learn RM structures from executions, yet assume that a labelling function is already available. 
The challenge is therefore to obtain both the task structure and its grounding in observations without relying on either extensive manual labelling or privileged access to the environment.

Vision-language models (VLMs) provide a potential source of visual reward supervision~\citep{baumli2023vision,rocamonde2023vision,ma2023liv,ma2024vision}. 
Directly using them to score non-Markovian tasks is difficult, however, because the significance of an observation depends on what preceded it. 
Moreover, VLM predictions over image sequences can be biased by temporal position: an observation may receive a higher score because it appears later in a sequence, rather than because the agent has made meaningful progress~\citep{ma2024vision}. This suggests a complementary division of labour: an RM can track progress through the task, while a VLM can identify what is happening in each image.

We introduce \themethodsparse{}, an approach that combines these capabilities for RM-guided policy learning from image-based observations. 
Given a natural-language task description, a large language model (LLM) generates an RM together with the atomic propositions that label its transitions. 
A VLM then grounds these propositions in observations, acting as the labelling function. 
The RM retains the history needed to interpret each observation, so the VLM can assess individual images without having to infer task progress from an ordered sequence.
Visual grounding is imperfect: an incorrect proposition label can cause an incorrect RM transition and degrade policy learning~\citep{nodari2023adversarial}. 
\themethodsparse{} therefore maintains an \emph{RM state belief}, a probability distribution over the agent's possible positions in the automaton, rather than committing to a single RM state. 
It defines reward through changes in expected accumulated RM reward under this belief. 
To supplement the rewards associated with RM transitions, we further extend \themethodsparse{} with Generative Value Learning (GVL), which estimates progress towards the next transition and provides a dense learning signal.
The resulting reward can be used with standard RL algorithms, both offline, reusing a single image dataset across tasks, and online, via a classifier distilled from the VLM labels.

We evaluate \themethodsparse{} on three domains spanning non-Markovian grid-world navigation, robotic manipulation, and partially observable 3D object collection. 
Across these domains, \themethodsparse{} consistently achieves higher task success rates than strong VLM-based reward baselines.

This paper makes the following contributions:
\begin{enumerate}
\item \textbf{Reward machines from natural language.} A novel method for deriving RMs from natural-language task descriptions together with the atomic propositions for labelling RM transitions. 
\item \textbf{VLMs as labelling functions.} Support for RM-guided policy learning from images, using a VLM to ground atomic propositions in image observations, both in the offline setting, where a single VLM-labelled image dataset is reused across tasks, and in the online setting, through a classifier distilled from the VLM labels.
\item \textbf{Robustness to imperfect grounding.} An RM state belief mechanism for arbitrary RM structures that defines a reward function based on changes in expected accumulated rewards over RM states, to limit the impact of incorrect grounding of RM transition labels.
\item \textbf{Dense rewards between RM transitions.} A GVL-based zero-shot estimate of progress towards the next RM transition, as a dense learning signal that supplements the RM sparse reward and accelerates policy learning without changing final performance.
\end{enumerate}

\section{Preliminaries}
\label{sec:preliminaries}
This section reviews the key concepts of reinforcement learning and reward machines. We use the following notation throughout the paper: for a set $\arbitraryset$, $\simplex{\arbitraryset}$ denotes the probability simplex over $\arbitraryset$ and $\powerset{\arbitraryset}$ its power set.

\subsection{Reinforcement learning}
\label{sec:prelim-rl}

Typically, a reinforcement learning process is formulated as a Markov Decision Process (MDP) 
$\mdp = \langle\states, \actions, \transitionprob, r, \discountfactor, \initialdist\rangle$, where $\states$ is a set of states, $\actions$ is a set of actions, $\transitionprob:\states\times\actions\to\simplex{\states}$ is a transition probability function, $r:\states\times\actions\times\states\to\mathbb{R}$ is a reward function, $\discountfactor\in[0,1)$ is the discount factor, and $\initialdist\in\simplex{\states}$ is the initial state distribution. At time $t$, the agent in state $s_t \in \states$ takes action $a_t \in \actions$, moves to $s_{t+1}\sim p(\cdot \mid s_t, a_t)$ and receives a reward $r_t = r(s_t,a_t,s_{t+1})$.
The goal of RL is to learn a policy $\pi$ that maximises the expected discounted return $\mathbb{E}\big[\sum_{t=0}^{\infty}\discountfactor^t r_t\big]$. The \emph{value function} $V^\pi(\statee) = \mathbb{E}_\pi\big[\sum_{t=0}^{\infty}\discountfactor^t r_t \mid \statee_0 = \statee\big]$ is the expected return from state $\statee$ under $\pi$. In \emph{online} RL, the policy is learned from interaction with the environment, and in \emph{offline} RL~\citep{levine2020offline}, from a fixed dataset $\dataset$ of trajectories, without any interaction. Sparse rewards make it more challenging for RL algorithms to learn a policy. To accelerate the learning, one can supplement the sparse reward with \emph{potential-based shaping}~\citep{ng1999policy}, which adds
$\discountfactor\Phi(\statee_{t})-\Phi(\statee_{t-1})$ to each reward
for a potential function $\Phi:\states\to\reals$, while leaving the
optimal policy unchanged.

\subsection{Reward machines}
\label{sec:prelim-rm}
Reward machines \citep[RM;][]{ToroIcarteKVM18,ToroIcarteKVM22} represent reward functions as finite-state machines that encode expressive, interpretable specifications of temporal task structure and use the RM states as external memory for non-Markovian tasks. RM transitions are labelled by subsets of $\propositions$, a set of atomic propositions abstracting the environment state space $\states$. A labelling function $\labelingfunction:\states\to\powerset{\propositions}$ maps each environment state to the set of propositions that hold in it, with $|\powerset{\propositions}|\ll|\states|$.
An RM is formally defined as follows.

\begin{definition}[Reward machine]
\label{def:reward-machine}
A \emph{Reward Machine} (RM) %
is a tuple $\rewardmachine=\langle\rmstates,\propositions,\rmstatetransitionfun,\rmrewardtransitionfun, \initialrmstate,\acceptingrmstate,\rejectingrmstate\rangle$, where $\rmstates$ is a finite set of states, $\propositions$ is a finite set of atomic propositions, $\rmstatetransitionfun:\rmstates\times\powerset{\propositions}\to\rmstates$ is the state-transition function, 
$\rmrewardtransitionfun:\rmstates\times\rmstates\to\reals$ is the reward-transition function such that 
$\rmrewardtransitionfun(u,u^{'})$ is the reward for transitioning from state $u$ to $u^{'}$. $\initialrmstate\in\rmstates$ is the initial state, and $\acceptingrmstate,\rejectingrmstate\in\rmstates$ are optional accepting and rejecting states.
\end{definition}

We write each transition as an edge $\langle\rmstate,\rmstate',\varphi,r\rangle$, where $\varphi$ is a propositional formula over $\propositions$ and $r=\rmrewardtransitionfun(\rmstate,\rmstate')$.

During interaction with the environment, the agent tracks its current RM state alongside its environment state.
Formally,
a label trace $\lambda=\langle \labelobs_0,\ldots,\labelobs_n\rangle$ induces a unique traversal $\rewardmachine(\lambda)=\langle \rmstate_0,\ldots,\rmstate_{n+1}\rangle$, where $\rmstate_{i+1}=\rmstatetransitionfun(\rmstate_i,\labelobs_i)$ and reward $\rmrewardtransitionfun(\rmstate_i,\rmstate_{i+1})$ is received at step $i$, and the episode ends when the traversal reaches $\acceptingrmstate$ or $\rejectingrmstate$. RM states summarise task history, which makes rewards that are non-Markovian over $\states$ Markovian over $\states\times\rmstates$.
We can thus learn a policy $\pi(a \mid \statee, \rmstate)$ over pairs of environment and RM states, with rewards given by the RM~\citep{ToroIcarteKVM22}.

\section{Methodology}
\label{sec:methodology}

We now describe RM-VLM, our approach for solving non-Markovian tasks from natural language descriptions and images without access to the environment's internal state. RM-VLM addresses three challenges: the automated generation of RM automata from natural language task descriptions, with transitions labelled by atomic propositions, the grounding of proposition labels in images, and the induction of a reward mechanism that provides dense feedback and remains robust to imperfect grounding. 
We first formalise the problem (\Cref{sec:problem-setting}) and describe how an LLM generates an RM from a natural-language task description and a VLM grounds its transition label propositions in images (\Cref{sec:grounded-rms}). We then explain how to estimate the RM state under label noise (\Cref{sec:rm-belief}) and complement this estimate with a dense reward signal to guide the RM-based policy learning (\Cref{sec:dense-reward}).

\subsection{Problem formulation}
\label{sec:problem-setting}
\begin{figure*}[t]
\centering
\includegraphics[width=\textwidth]{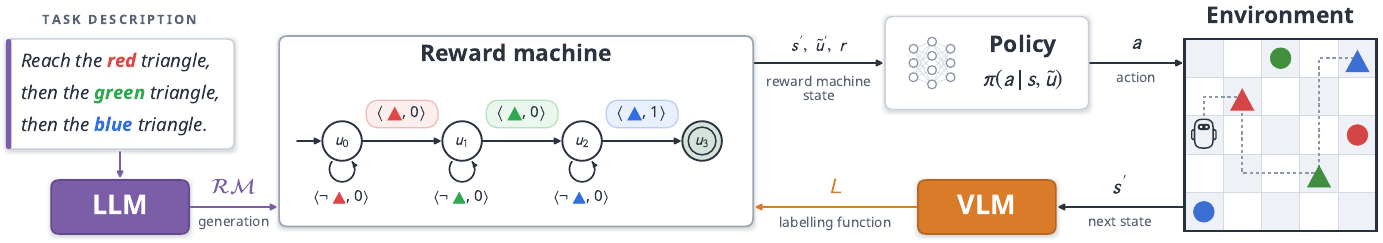}
\caption{The reinforcement learning loop considered in this work. An LLM compiles the language command into reward machines over atomic propositions, and a VLM grounds those propositions in image observations to drive RM transitions. The reward obtained from the LLM-generated RM is used to supervise policy learning.}
\label{fig:method-overview}
\end{figure*}

We consider the case where an agent does not have access to the RM of a task, its propositions, or the labelling function. We assume a given dataset $\dataset = \{\trajectory_j\}_{j=1}^{M}$ of $M$ trajectories $\trajectory_j = \langle \statee_0, a_0, \ldots, \statee_T \rangle$, collected by a behaviour policy in the environment without rewards or labels. Environment states are observed as images, and tasks are given as natural-language descriptions $\rmtext{i} \in \taskset$. Given a natural-language task description $\rmtext{i}$, an RM-VLM reward machine is a tuple $\llmrewardmachine$ whose atomic propositions in $\propositions$ are associated with the task description $\rmtext{i}$. The propositions and the state transition function of $\llmrewardmachine$ are generated by an LLM from the task description augmented with a natural-language description of the environment and its physical constraints as described in Section~\ref{sec:grounded-rms}. The agent tracks its current state, by grounding the label propositions of the RM transitions in the image-based environment state, using the (noisy) labelling function $\noisylabelingfunction:\mathcal{I}\to\powerset{\propositions}$, where $\mathcal{I}$ is the space of image observations, which returns the propositions predicted to hold in the image of an environment state. \Cref{fig:method-overview} presents an overview of our RM-VLM approach. An LLM  first generates a set of propositions $\llmpropositions$ from the natural-language task descriptions, then generates an $\llmrewardmachine$ for each $\rmtext{}$ using these propositions. At each step, a VLM acts as a noisy labelling function $\noisylabelingfunction$, predicting which propositions are true in the image-based environment state. These labels are used to compute the RM's state belief and the reward, guiding the agent to learn a policy that maximises the expected return.

\subsection{VLM-grounded reward machines from language}
\label{sec:grounded-rms}
RM-VLM uses a large language model (LLM) to generate the components of an RM that provides an automaton representation of a natural-language task description. Specifically, given a set $\{\rmtext{i}\}_{i=1}^N$ of task descriptions, an LLM is prompted to generate the set $\llmpropositions$ of propositions shared across all tasks. Then, for each task $\rmtext{i}$ an associated $\llmrewardmachine_i$ is generated over the alphabet $\llmpropositions$. This is done by prompting the LLM to decompose each $\rmtext{i}$ into a set of independent sub-task RMs, which are then combined together into a single $\llmrewardmachine_i$~\citep{neary2020reward}.
We evaluate the reliability of proposition and RM generation in \Cref{sec:rm-construction}.
Given that generating RMs requires only a few calls, a high-quality LLM can be used for this step.

To ground the propositions, we query a VLM for each episode in $\dataset$ and each proposition $\proposition \in \llmpropositions$ to identify the timesteps at which $\proposition$ holds, producing a labelled dataset $\labelleddataset$.
Since its predictions can be wrong, we treat the VLM as a noisy labelling function $\noisylabelingfunction$ (\Cref{sec:problem-setting}).

We describe how $\labelleddataset$ is used for offline and online policy learning in \Cref{sec:dense-reward}, and include the prompts used for proposition extraction and RM generation in \Cref{app:prompts}.
\subsection{Reward machine state belief}
\label{sec:rm-belief}

In RM-based reinforcement learning, the environment state is typically augmented with the current RM state (\Cref{sec:prelim-rm}).
When proposition labels are provided by a perfect labelling function, the current RM state can be tracked deterministically (\Cref{sec:prelim-rm}).
With VLM-derived (noisy) labels, however, deterministic state tracking can cause significant degradation in policy performance~\citep{nodari2023adversarial}.
To make tracking robust, we replace the deterministic state tracking with an \emph{RM state belief}, a probability distribution over RM states updated online at each timestep~\citep{li2024reward,parac2024learning}.
Prior schemes such as \emph{Independent Belief Updating (IBU)}~\citep{li2024reward} re-weight transitions by the predicted labels alone.
We additionally exploit two sources of structure.
First, given the environment's physical constraints, described in natural language (\Cref{lst:env-constraints}), the generated RMs encode causal preconditions between propositions, even when they are not explicitly required by the task.
For instance, if a drawer must be unlocked before it can be opened, a false positive on ``drawer open'' cannot trigger a premature transition to the accepting state.
Second, a state-dependent prior over true proposition labels couples independently predicted propositions by encoding structural co-occurrence constraints between them.
For example, if every object has both a colour and a shape, the prior can encode that a shape proposition never holds without a colour proposition, reducing the impact of isolated misdetections.
We formalise the belief update below with the help of a \emph{transition--emission matrix}.

\begin{definition}[Transition--emission matrix]
\label{def:transition-emission}
Let $\noisylabelobs \subseteq \propositions$ be a label predicted by a noisy labelling function,
$\labelobs \subseteq \propositions$ be a label that truly holds, and $\rmstate, \rmstate' \in \rmstates$ a pair of RM states.
The transition--emission matrix $C \in \mathbb{R}^{|\rmstates| \times |\rmstates| \times 2^{|\propositions|}}$ is defined as the joint probability of transitioning from state $\rmstate$ to $\rmstate'$ and observing the noisy label $\noisylabelobs$:

\begin{equation}
\label{eq:C-matrix-main}
C[\rmstate, \rmstate', \noisylabelobs] = \sum_{\labelobs \in \powerset{\propositions}} P(\labelobs \mid \rmstate) \cdot \mathbf{1}\!\big[\rmstatetransitionfun(\rmstate, \labelobs) = \rmstate'\big] \cdot P(\noisylabelobs \mid \labelobs),
\end{equation}
where $P(\labelobs \mid \rmstate)$ is the state-dependent label prior and $P(\noisylabelobs \mid \labelobs)$ is the probability of observing a noisy label given the ground-truth label, which factorises as $P(\noisylabelobs \mid \labelobs) = \prod_{\proposition \in \propositions} P(\noisylabelobs_\proposition \mid \labelobs_\proposition)$ with
\begin{equation}
\label{eq:obs-noise}
P(\noisylabelobs_\proposition = 1 \mid \labelobs_\proposition) = \labelobs_\proposition\,\mathrm{TPR}_\proposition + (1 - \labelobs_\proposition)\,\mathrm{FPR}_\proposition,
\end{equation}
where $\noisylabelobs_\proposition$ (resp. $\labelobs_\proposition$) refers to proposition $\proposition$ in $\noisylabelobs$ (resp. in $\labelobs$), $P(\noisylabelobs_\proposition = 0\mid \labelobs_\proposition) = 1 - P(\noisylabelobs_\proposition = 1\mid \labelobs_\proposition)$, and $\mathrm{TPR}_\proposition$ and $\mathrm{FPR}_\proposition$ are the true and false positive rates of the noisy labelling function for proposition $\proposition$.
\end{definition}

In practice, we estimate $\mathrm{TPR}_\proposition$ and $\mathrm{FPR}_\proposition$ by comparing the predicted labels with ground-truth labels on a set of trajectories from $\dataset$.
We also generate the prior $P(\labelobs \mid \rmstate)$ with the LLM from the environment description and the generated RM (see \Cref{app:rm-construction}).

\begin{definition}[RM state belief]
\label{def:rm-belief}
The RM state belief $\rmbelief_t \in \simplex{\rmstates}$ is a probability distribution over RM states at timestep $t$, defined as:
\begin{align}
\hat{\rmbelief}_{t}[\rmstate'] &= \begin{cases}
\mathbf{1}[\rmstate' = \initialrmstate] & \text{if } t = 0, \\[4pt]
\sum_{\rmstate} \rmbelief_{t-1}[\rmstate] \cdot C[\rmstate, \rmstate', \noisylabelobs_{t-1}] & \text{if } t > 0,
\end{cases} \label{eq:belief-update} \\[4pt]
\rmbelief_{t} &= \frac{\hat{\rmbelief}_{t}}{\sum_{\rmstate'} \hat{\rmbelief}_{t}[\rmstate']}, \nonumber
\end{align}
where $\initialrmstate$ is the initial RM state, $\noisylabelobs_{t-1}$ is the noisy label of the image for the environment state $\statee_{t-1}$, and $C$ is the transition--emission matrix.
\end{definition}

An illustrative example is given in \Cref{app:belief-grid-example}.

\subsection{Reward and policy learning}
\label{sec:dense-reward}
We present a novel way of computing RM rewards based on the notion of \emph{expected accumulated reward}.
First, let $G_t = \sum_{k=1}^{t} \rmrewardtransitionfun(\rmstate_{k-1}, \rmstate_k)$ denote the ground-truth reward the RM has emitted up to time $t$.
Also, let $\rmalpha_t[\rmstate] = \mathbb{E}\big[G_t \mid \rmstate_t = \rmstate, \noisylabelobs_{0:t-1}\big]$ be its expectation given the observed label history.
We initialise $\rmalpha_0 = \mathbf{0}$ and recursively compute $\rmalpha$ for each RM state $\rmstate'$ as:
\begin{equation}
\label{eq:alpha-update}
\rmalpha_{t}[\rmstate'] =
\frac{\sum_{\rmstate} \rmbelief_{t-1}[\rmstate]\, C[\rmstate, \rmstate', \noisylabelobs_{t-1}] \,\big( \rmalpha_{t-1}[\rmstate] + \rmrewardtransitionfun(\rmstate, \rmstate') \big)}
     {\sum_{\rmstate} \rmbelief_{t-1}[\rmstate]\, C[\rmstate, \rmstate', \noisylabelobs_{t-1}]},
\end{equation}
where $C$ is the transition-emission matrix of RM. The task reward $\delta_r(\rmbelief_{t-1}, \rmbelief_t)$ at time $t$ is defined as:
\begin{equation}
\label{eq:task-reward}
\delta_r(\rmbelief_{t-1}, \rmbelief_t) = (\rmbelief_t)^\top \rmalpha_t - (\rmbelief_{t-1})^\top \rmalpha_{t-1}.
\end{equation}
This task reward reduces exactly to the RM reward when the RM state belief is one-hot (\Cref{app:one-hot-proof}).

We augment the task reward with a dense reward signal term $\densereward_t$, which builds upon two existing shaping schemes: rewarding progress across RM transitions \citep{camacho2019ltl} and towards them \citep{li2025ground}.
Both are potential-based shaping terms (\Cref{sec:prelim-rl}) built from a value function over RM states.
We first compute the value of each RM state by value iteration over the RM automaton alone:
\begin{equation}
\label{eq:rm-vi}
\rmpotential(\rmstate) = \max_{\langle \rmstate, \rmstate', \varphi, r \rangle}
  \rmdiscount \big( r + \rmpotential(\rmstate') \big),
\end{equation}
where the maximum is taken over all outgoing transitions $\langle \rmstate, \rmstate', \varphi, r \rangle$ of $\rmstate$, terminal states are fixed to $\rmpotential(\rmstate) = 0$, and $\rmdiscount$ is a discount factor over RM transitions.
As the value iteration algorithm considers RM transitions to be the singular actions, it should be run with
$\rmdiscount < \discountfactor$ as it takes many low-level steps to achieve a transition.
We then extend this value function to include also the environment states ($\rmvalue: \states \times \rmstates \rightarrow \reals$), thus estimating how close the agent is to triggering an RM transition. This estimation is computed as follows. GVL~\citep{ma2024vision} is used to zero-shot estimate  the value of reaching each atomic proposition $\vevt{\proposition}(\statee)$ and its negation $\vevt{\neg\proposition}(\statee)$ (\Cref{app:gvl-progress}).
These estimates are composed according to the logical structure of $\varphi$, as in \citet{li2025ground}:
\begin{align}
\vevt{\varphi_1 \land \varphi_2}(\statee) &= \min\!\big(\vevt{\varphi_1}(\statee),\; \vevt{\varphi_2}(\statee)\big), \label{eq:evt-and} \\
\vevt{\varphi_1 \lor \varphi_2}(\statee) &= \max\!\big(\vevt{\varphi_1}(\statee),\; \vevt{\varphi_2}(\statee)\big). \label{eq:evt-or}
\end{align}
By design, $\vevt{\varphi}(\statee) \in [0,1]$ for any formula $\varphi$, as it is the value of reaching a state that satisfies $\varphi$ under a sparse reward of 1.
The full value estimate $\rmvalue(\statee, \rmstate)$, \themethoddense{} is given by combining $\vevt{\varphi}(\statee)$ with the RM state values $\rmpotential(\rmstate)$ and $\rmpotential(\rmstate')$ as follows:
\begin{align}
\rmvalue(\statee, \rmstate)
  &= \max_{\langle \rmstate, \rmstate', \varphi, r \rangle}
  \Big[ \big(1 - \vevt{\varphi}(\statee)\big) \cdot \rmpotential(\rmstate) \nonumber \\
  &\qquad + \vevt{\varphi}(\statee) \cdot \big(r + \discountfactor \cdot \rmpotential(\rmstate') \big) \Big]. \label{eq:rm-value}
\end{align}
Finally, $\rmvalue$ is applied to the RM state belief by taking the expectation under the belief:
\begin{equation}
\label{eq:belief-value}
\rmvalue(\statee,\rmbelief)
= \sum_{\rmstate \in \rmstates} \rmbelief[\rmstate]\, \rmvalue(\statee,\rmstate).
\end{equation}

The dense reward signal term is then given by: 
\begin{equation}
\label{eq:belief-reward}
\densereward_t
  = \discountfactor\, \rmvalue(\statee_{t}, \rmbelief_{t}) - \rmvalue(\statee_{t-1}, \rmbelief_{t-1}).
\end{equation}
The total reward used for policy learning is given by the value $r_t = \delta_r(\rmbelief_{t-1}, \rmbelief_t) + \densereward_t$.

\paragraph{Policy learning.}
The tasks we consider can have non-Markovian rewards. 
We therefore augment the environment state with the RM state belief from \Cref{def:rm-belief} and use $r_t$ as the reward signal.

We consider both offline and online RL settings.
In the offline setting, the RM state beliefs and rewards can be computed from labelled offline trajectories without further environment interaction. We use the same dataset to learn policies for different tasks by augmenting it with  different self-generated rewards. In the online setting, since querying the VLM at every step is too costly, we distil the VLM labels into a neural network classifier trained to label states, and use it during online RL.

\section{Experiments}
\label{sec:experiments}

\begin{table*}[t]
\centering
\caption{Compositional language tasks evaluated in each environment.}
\label{tab:tasks}
\begin{tabular}{llp{0.66\textwidth}}
\toprule
\textbf{Env} & \textbf{Task} & \textbf{Language command} \\
\midrule
\multirow{4}{*}{GeoGrid}
 & Sequence & Go to a red triangle, then a green triangle, then a blue triangle. \\
 & Loop     & Repeatedly go to a red triangle, then a green triangle, then a blue triangle. \\
 & Logic    & Go to all six objects, but always go to red objects before blue objects, and blue objects before green objects. \\
 & Safety   & Go to a red object, then a blue object, then a green object, but always avoid triangles. \\
\midrule
\multirow{4}{*}{OGBench-Scene}
 & Open & Open both the drawer and the window. \\
 & Close+Lock & Close the open drawer and window, then re-engage both locks. \\
 & Cube-in-Drawer & Place the cube into the drawer. \\
 & Cube+Window & Place the cube into the drawer and open the window. \\
\midrule
\multirow{4}{*}{MiniWorld-Pickup}
 & Sequence    & Pick up the red box, then the green box, then the blue box. \\
 & Logic       & Pick up all six objects, but always pick up red objects before blue objects, and blue objects before green objects. \\
 & Safety      & Pick up the red ball, then the blue ball, then the green ball, but never pick up a box. \\
 & Alternation & Pick up a ball, then a box, then another ball, in that order. \\
\bottomrule
\end{tabular}
\end{table*}

To evaluate the effectiveness of our methods, we design experiments around three research questions:

\begin{sloppypar}
\begin{description}
  \item[RQ1 --- Effectiveness.]
   Can \themethodsparse{} solve non-Markovian tasks without a pre-engineered labelling function and task-specific RM?

  \item[RQ2 --- Grounding decomposition with RMs.] Does decomposing a task into an RM yield better reward and faster learning than using the VLM directly?

  \item[RQ3 --- Robustness to grounding noise.]
   How do VLM labelling errors impact policy learning?

\end{description}
\end{sloppypar}

We describe the experimental setup in \Cref{sec:environments}, evaluate the LLM-based generation of RM in \Cref{sec:rm-construction}, present the offline RL results in \Cref{sec:main-results} and the online RL results in \Cref{sec:online-rl}, and present an ablation study of our design choices in \Cref{sec:ablations}.

\subsection{Experimental setup}
\label{sec:environments}
We evaluate RM-VLM on three different domains: \emph{GeoGrid}~\citep{li2025ground}, \emph{OGBench-Scene}~\citep{park2024ogbench}, and \emph{MiniWorld-Pickup}~\citep{chevalier2023minigrid}.
\emph{GeoGrid} is an $8\times8$ grid world where the agent visits specified objects among three triangles and three circles coloured red, green, or blue, moving along four cardinal directions.
Its simple visual setting provides a clean setting for evaluating the generated reward.
\emph{OGBench-Scene} is a continuous-control robotic manipulation domain that requires the agent to complete manipulation tasks involving a drawer, a window, two lock buttons, and a cube.  Its rendered robot scenes are substantially
harder for the VLM to label than GeoGrid’s symbolic grids, so it tests robustness of RM-VLM to inaccurate VLM labels and long-horizon control from images. \emph{MiniWorld-Pickup} is a goal-oriented 3D environment requiring the agent to collect specified objects among three balls and three boxes coloured red, green, or blue. Its first-person 3D observations test performance under partial observability. \Cref{tab:tasks} lists the RM tasks we consider in each of these domains, requiring different reward structures: fixed sequence of sub-goals, repeated objectives, ordering constraints between sub-goals, and safety constraints. For each task, policy learning uses expert-written reference propositions, RMs and label priors. \Cref{sec:rm-construction} shows the LLM generation of these from the task descriptions, whereas further details including VLM labelling accuracies, and descriptions of offline datasets are included in \Cref{app:env_descriptions}.

\paragraph{Baselines.}
For offline RL, we compare \themethodsparse{} against three baselines that do not use reward machines.
\textbf{Behaviour Cloning (BC)}~\citep{pomerleau1988alvinn} imitates the behaviour policy used to collect the dataset.
\textbf{Generative Value Learning (GVL)}~\citep{ma2024vision} uses a VLM to estimate task progress from shuffled frames and uses this estimate as a reward.
\textbf{VLM Success Detection (VLM-SD)}~\citep{baumli2023vision} uses a VLM to judge whether the task has been completed and assigns a sparse success reward.
Both VLM baselines receive the full language command from \Cref{tab:tasks} as task description (prompts in \Cref{app:prompts}), whereas our methods query the VLM only about individual sub-goals.
Since the baselines have no RM state to track sub-goal completion, they use a recurrent policy on GeoGrid and MiniWorld-Pickup, where completed sub-goals are not visible in the observation.
In the online RL setting, we compare against \textbf{Ground-Compose-Reinforce (GCR)}~\citep{li2025ground}, which uses reward machines but assumes that $\labelleddataset$ is labelled by an oracle, therefore tracking RM state deterministically.
\textbf{GCR-High-level RS} provides reward on transitions between RM states, whereas \textbf{GCR} also rewards progress towards the next RM transition, estimated by offline RL learned value functions on $\labelleddataset$.

We use Gemma-4-31B~\citep{team2026gemma4} as both LLM and VLM. In the offline setting, we train policies with Implicit Q-Learning~\citep[IQL;][]{kostrikov2021offline} and, on OGBench-Scene, also with Flow Q-Learning~\citep[FQL;][]{park2025fql}, specific for continuous action spaces. In the online setting, we use Proximal Policy Optimization~\citep[PPO;][]{schulman2017proximal}.
\themethodsparse{} trains a feed-forward policy on image observations extended with RM state belief.
We report two variants: \textbf{\themethodsparse{}}, which uses the task reward and RM-based shaping, and \textbf{\themethoddense{}}, which additionally uses the GVL progress estimate (\Cref{sec:dense-reward}).
Full training details are given in \Cref{app:hyperparams}.

\subsection{Evaluating Proposition and RM generation}
\label{sec:rm-construction}

Errors in the generation of propositions or RMs would affect policy learning (RQ1). We therefore evaluate each step against the reference propositions, RMs and label priors, over 10 runs per task. Firstly, the LLM generates the propositions from all task descriptions of an environment; results show that in every run, the LLM captures all reference propositions the tasks require.
The LLM then generates an RM over the generated propositions. Results show that the generated RM is trace-equivalent to the reference RM in $114/120$ runs. The remaining failures are malformed RMs.
Given the reference propositions instead, it does so in every run ($120/120$), in line with prior work~\citep{li2025ground,castanyer2025arm}.
Given the reference RMs, the generated label prior matches the reference prior in $119/120$ runs, and the remaining error comes from the LLM's failure to follow the defined structure.
We report the full results in \Cref{app:rm-construction}.

\begin{table*}[t]
\centering
\small
\setlength{\tabcolsep}{4pt}
\caption{Task success rates with IQL~\citep{kostrikov2021offline} (mean $\pm$ std over 5 seeds). Values within $95\%$ of the best per row in bold.}
\label{tab:main_results}
\begin{tabular}{llccccc}
\toprule
\textbf{Env} & \textbf{Task} & \textbf{BC} & \textbf{GVL} & \textbf{VLM-SD} & \textbf{\themethodsparse} & \textbf{\themethoddense} \\
\midrule
\multirow{4}{*}{GeoGrid}
 & Sequence & $0.22 \pm 0.01$ & $0.31 \pm 0.09$ & $0.14 \pm 0.02$ & $\mathbf{1.00 \pm 0.00}$ & $\mathbf{1.00 \pm 0.00}$ \\
 & Loop$^\dagger$ & $0.34 \pm 0.01$ & $0.21 \pm 0.06$ & $0.09 \pm 0.02$ & $\mathbf{1.00 \pm 0.00}$ & $\mathbf{1.00 \pm 0.00}$ \\
 & Logic    & $0.00 \pm 0.00$ & $0.00 \pm 0.01$ & $0.00 \pm 0.00$ & $\mathbf{0.98 \pm 0.01}$ & $\mathbf{0.97 \pm 0.01}$ \\
 & Safety   & $0.10 \pm 0.01$ & $0.09 \pm 0.03$ & $0.15 \pm 0.02$ & $\mathbf{0.95 \pm 0.01}$ & $\mathbf{0.95 \pm 0.01}$ \\
\midrule
\multirow{4}{*}{OGBench-Scene}
 & Open & $0.60 \pm 0.05$ & $0.63 \pm 0.15$ & $0.73 \pm 0.15$ & $\mathbf{0.98 \pm 0.02}$ & $\mathbf{0.97 \pm 0.03}$ \\
 & Close+Lock & $0.09 \pm 0.03$ & $0.10 \pm 0.04$ & $0.09 \pm 0.07$ & $0.53 \pm 0.18$ & $\mathbf{0.58 \pm 0.20}$ \\
 & Cube-in-Drawer & $0.04 \pm 0.02$ & $0.06 \pm 0.04$ & $0.02 \pm 0.03$ & $\mathbf{0.87 \pm 0.10}$ & $0.80 \pm 0.10$ \\
 & Cube+Window & $0.01 \pm 0.01$ & $0.02 \pm 0.02$ & $0.00 \pm 0.00$ & $\mathbf{0.73 \pm 0.04}$ & $0.65 \pm 0.11$ \\
\midrule
\multirow{4}{*}{MiniWorld-Pickup}
 & Sequence    & $0.15 \pm 0.03$ & $0.21 \pm 0.04$ & $0.20 \pm 0.06$ & $\mathbf{0.74 \pm 0.08}$ & $0.67 \pm 0.05$ \\
 & Logic       & $0.25 \pm 0.05$ & $0.15 \pm 0.03$ & $0.25 \pm 0.02$ & $\mathbf{0.74 \pm 0.07}$ & $\mathbf{0.72 \pm 0.06}$ \\
 & Safety      & $0.16 \pm 0.02$ & $0.18 \pm 0.04$ & $0.18 \pm 0.02$ & $\mathbf{0.56 \pm 0.03}$ & $0.50 \pm 0.05$ \\
 & Alternation & $0.37 \pm 0.04$ & $0.38 \pm 0.06$ & $0.44 \pm 0.06$ & $\mathbf{0.93 \pm 0.02}$ & $\mathbf{0.93 \pm 0.03}$ \\
\bottomrule
\end{tabular}

\smallskip
{\footnotesize $^\dagger$Completed cycles per episode, normalised by the ground-truth oracle's cycle count.}
\end{table*}

\subsection{Compositional-task evaluation}
\label{sec:main-results}

\Cref{tab:main_results} reports task success of \themethodsparse{} and our baselines with IQL on GeoGrid, OGBench-Scene, and MiniWorld-Pickup.
In line with \citet{jackson2026clean}, we select the best performing checkpoint online with a fixed budget and report its performance on separate held-out episodes (\Cref{app:hyperparams}).

Overall, \themethodsparse{} solves the tasks
effectively across the three domains (RQ1). It reaches perfect or near-perfect success rate on all GeoGrid tasks, and solves three of four OGBench-Scene tasks with success rate above $0.65$.
\themethodsparse{} outperforms all non-RM baselines on every task across the three environments (RQ2), suggesting that grounding atomic propositions and composing them with an RM is more effective than asking the VLM to judge full-task progress or success directly.
Finally, incorporating GVL does not have a substantial impact on the final performance of the agent.
Repeating the OGBench-Scene evaluation with FQL (\Cref{app:scene-fql}) shows that the gains are not specific to IQL.

\subsection{Online RL on GeoGrid}
\label{sec:online-rl}

In the online setting, we train PPO~\citep{schulman2017proximal} from scratch on GeoGrid with \themethodsparse{} and \themethoddense{}, and compare them against GCR-High-level RS and GCR.
All four methods distil a proposition classifier from the same fixed dataset and use it as the labelling function during policy training.
The GCR variants distil it from ground-truth labels, and GCR additionally learns its value functions with offline RL on the same dataset.
Our variants distil it from VLM labels, and \themethoddense{} additionally distils its progress model from GVL predictions.

\begin{figure*}[t]
  \centering
  \includegraphics[width=\textwidth]{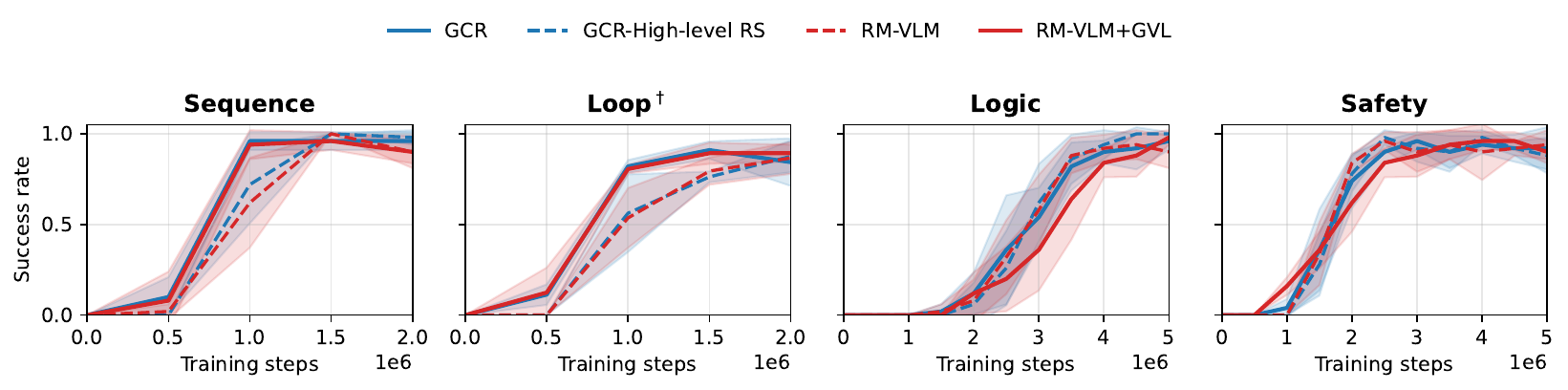}
  \caption{\textbf{Online RL on GeoGrid.} Ground-truth success rate during PPO training (mean $\pm$ std over 5 seeds).}
  \label{fig:online_rl_geogrid}
\end{figure*}

\Cref{fig:online_rl_geogrid} shows that \themethodsparse{} matches GCR-High-level RS and \themethoddense{} matches GCR on every task, showing that replacing ground-truth labels and learned value functions with VLM outputs incurs no loss in final performance on GeoGrid.

\subsection{Ablations}
\label{sec:ablations}

We present ablation study on RM state tracking,  reward signal and label quality, and analyse why RM-based decomposition helps.

\subsubsection{RM state belief and labelling errors}

To analyse the effect of RM-state tracking under VLM labels, \Cref{tab:ablation_belief} compares three variants:
\begin{itemize}
\item \emph{Naive}: update the RM state directly from VLM labels, treating them as correct \citep{li2024reward}.
\item \emph{Independent Belief Updating (IBU)}: maintain a belief over RM states using the predicted labels, as in \citet{li2024reward} and Prob-IRM~\citep{parac2024learning}, but without state-dependent structural prior.
\item \emph{Belief}: use the full belief update, including the RM structure and structural prior.
\end{itemize}
As an upper bound, it also reports an \emph{Oracle} variant of \themethodsparse{} that uses the same RM and reward construction but replaces VLM labels with ground-truth proposition labels, which measures the cost of imperfect VLM grounding.

\begin{table}[t]
\centering
\footnotesize
\setlength{\tabcolsep}{6pt}
\caption{Belief ablation: mean $\pm$ standard deviation of task success over 5 seeds. Impact of labelling errors: \textbf{\themethodsparse} with VLM labels vs.\ ground-truth labels (\textbf{Oracle}). Values within $95\%$ of the best per row in bold.}
\label{tab:ablation_belief}
\begin{tabular}{lcccc}
\toprule
\textbf{Task} & \textbf{Naive} & \textbf{IBU} & \textbf{Belief} & \textbf{Oracle} \\
\midrule
\multicolumn{5}{l}{\emph{GeoGrid}} \\
Sequence & $\mathbf{1.00 \pm 0.01}$ & $\mathbf{1.00 \pm 0.00}$ & $\mathbf{1.00 \pm 0.00}$ & $\mathbf{1.00 \pm 0.00}$ \\
Loop$^\dagger$ & $\mathbf{1.00 \pm 0.00}$ & $\mathbf{0.98 \pm 0.01}$ & $\mathbf{1.00 \pm 0.00}$ & $\mathbf{1.00 \pm 0.00}$ \\
Logic    & $\mathbf{0.99 \pm 0.00}$ & $\mathbf{0.97 \pm 0.02}$ & $\mathbf{0.98 \pm 0.01}$ & $\mathbf{0.99 \pm 0.01}$ \\
Safety   & $\mathbf{0.95 \pm 0.01}$ & $\mathbf{0.95 \pm 0.00}$ & $\mathbf{0.95 \pm 0.01}$ & $\mathbf{0.95 \pm 0.01}$ \\
\midrule
\multicolumn{5}{l}{\emph{OGBench-Scene}} \\
Open & $0.59 \pm 0.05$ & $0.49 \pm 0.17$ & $\mathbf{0.98 \pm 0.02}$ & $\mathbf{0.98 \pm 0.01}$ \\
Close+Lock & $0.16 \pm 0.08$ & $0.13 \pm 0.03$ & $0.53 \pm 0.18$ & $\mathbf{0.97 \pm 0.02}$ \\
Cube-in-Drawer & $\mathbf{0.84 \pm 0.02}$ & $\mathbf{0.84 \pm 0.04}$ & $\mathbf{0.87 \pm 0.10}$ & $\mathbf{0.86 \pm 0.06}$ \\
Cube+Window & $0.53 \pm 0.12$ & $0.44 \pm 0.06$ & $\mathbf{0.73 \pm 0.04}$ & $\mathbf{0.72 \pm 0.08}$ \\
\midrule
\multicolumn{5}{l}{\emph{MiniWorld-Pickup}} \\
Sequence & $\mathbf{0.77 \pm 0.06}$ & $\mathbf{0.77 \pm 0.04}$ & $\mathbf{0.74 \pm 0.08}$ & $\mathbf{0.75 \pm 0.03}$ \\
Logic & $\mathbf{0.71 \pm 0.03}$ & $\mathbf{0.73 \pm 0.03}$ & $\mathbf{0.74 \pm 0.07}$ & $\mathbf{0.74 \pm 0.05}$ \\
Safety & $\mathbf{0.55 \pm 0.04}$ & $0.53 \pm 0.04$ & $\mathbf{0.56 \pm 0.03}$ & $\mathbf{0.58 \pm 0.07}$ \\
Alternation & $\mathbf{0.93 \pm 0.02}$ & $\mathbf{0.93 \pm 0.02}$ & $\mathbf{0.93 \pm 0.02}$ & $\mathbf{0.93 \pm 0.02}$ \\
\bottomrule
\end{tabular}

\smallskip
{\footnotesize $^\dagger$Normalised by the Oracle's cycle count.}
\end{table}

The three approaches, unsurprisingly, achieve similar success rates in GeoGrid and MiniWorld-Pickup where the labels are reliable.
However, in OGBench-Scene, the \emph{Belief} shows its benefit, outperforming the other two variants in three out of four tasks and matching them on Cube-in-Drawer.
Under synthetic label noise on GeoGrid (\Cref{app:label-noise}), \emph{Belief} also degrades most gracefully, retaining $0.79$ and $0.85$ success on Logic and Safety at a $2\%$ flip rate where \emph{Naive} drops to $0.55$ and $0.70$.
On GeoGrid and MiniWorld-Pickup, VLM labels are accurate enough that \emph{Belief} matches the \emph{Oracle} on every task.
On OGBench-Scene, it matches the \emph{Oracle} on every task except Close+Lock ($0.53$ vs.\ $0.97$), the only task that requires detecting a closed drawer and window, which the VLM labels least reliably (RQ3).

\subsubsection{Reward signal}
\label{sec:reward-signal-ablation}

\begin{figure*}[t]
  \centering
  \includegraphics[width=\textwidth]{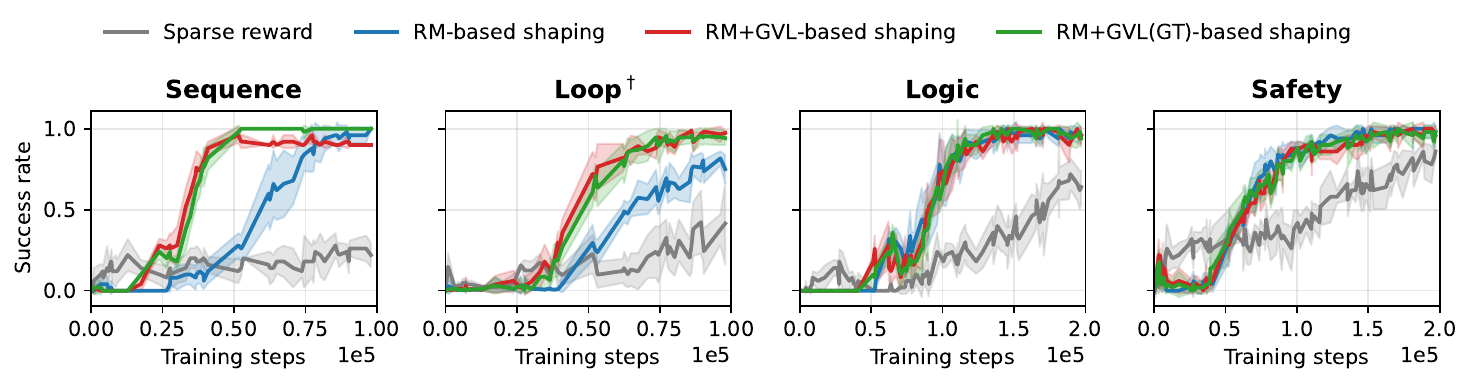}
  \caption{GeoGrid convergence curves for the reward-signal variants (mean $\pm$ std over 5 seeds).}
  \label{fig:gvl-rm-convergence}
\end{figure*}

We next ablate the components of the RM-based reward on GeoGrid, where VLM proposition labels are highly accurate.
We compare four variants:
\begin{itemize}
  \item \emph{sparse reward}: use only the sparse reward from the RM transition function $\rmrewardtransitionfun$.
  \item \emph{RM-based shaping}: adds RM-based shaping obtained by value iteration described in \Cref{eq:rm-vi}.
  \item \emph{RM+GVL-based shaping}: additionally predicts progress towards reaching proposition with GVL, combining its predictions into a value estimate described in \Cref{eq:rm-value}.
  \item \emph{RM+GVL(GT)-based shaping}: same as above but replacing the GVL-progress prediction with the ground truth one.
\end{itemize}
\Cref{fig:gvl-rm-convergence} shows the resulting convergence curves.

\emph{RM-based shaping} improves convergence over the \emph{sparse reward}, showing that RM structure can provide useful dense guidance.
\emph{RM+GVL-based shaping} can further accelerate learning at early stages when the progress signal is accurate, but the additional VLM noise it introduces can hurt final performance, as seen on Sequence (RQ3).
On Logic and Safety, we do not see the same benefits.
The issue arises because the use of min~\citep[Appendix~A.4]{li2025ground} overestimates progress towards the next transition, which in turn can occasionally generate a misleading reward signal.

\subsubsection{Why does RM-based decomposition help?}
\label{sec:rm-benefits}

GeoGrid, despite being a simple environment, poses significant challenges to GVL (RQ2).
Focusing on value function estimation, we examine two capabilities that RMs provide, task decomposition and memory.
We design tasks such that only these capabilities are individually evaluated.
We compare the value function used by the \themethoddense{} densification term (\Cref{sec:dense-reward}) against the one produced by GVL, measuring \emph{Value-Order Correlation (VOC)}~\citep{ma2024vision}, the rank correlation between the predicted values and the timestep order along an optimal trajectory, for simple GeoGrid tasks.

\emph{Task decomposition.}
A VLM answering a disjunctive query must jointly reason about all alternatives simultaneously, which could degrade progress estimation.
We compare GVL with a disjunctive prompt against decomposing the task into atomic sub-goal queries and taking their per-frame maximum, following fuzzy-logic disjunction~\citep{zadeh1965fuzzy}.
For the disjunctive task ``reach a red object OR a triangle'', we collect 200 optimal episodes (100 per sub-goal) and compute progress with both approaches.
\Cref{tab:voc} reports the results.

\begin{table}[t]
\centering
\small
\caption{VOC of GVL under conditions that isolate task decomposition and memory.
Higher is better; a perfect value estimator achieves $+1.0$.}
\label{tab:voc}
\begin{tabular}{lc}
\toprule
\textbf{Condition} & \textbf{VOC (mean $\pm$ SE)} \\
\midrule
\multicolumn{2}{l}{\emph{Decomposition} (``reach red OR triangle'')} \\
Disjunctive prompt     & $+0.931 \pm 0.010$ \\
Per-atom queries (max) & $+0.944 \pm 0.010$ \\
\midrule
\multicolumn{2}{l}{\emph{Memory} (``reach triangle, then circle'')} \\
Standard env                & $+0.541 \pm 0.028$ \\
Yellow-marker env           & $+0.878 \pm 0.006$ \\
VLM sub-goal split          & $+0.986 \pm 0.002$ \\
\bottomrule
\end{tabular}
\end{table}

Per-atom queries perform on par with the disjunctive prompt, so decomposing the query does not lose accuracy.
The RM decomposes tasks into such atomic queries by construction (\Cref{sec:dense-reward}).

\emph{Memory.}
Sequential tasks require tracking sub-goal completion across time, but GVL presents frames in random order, so the VLM must rely solely on the observation to infer this history.
We compare GVL against two alternatives on the task ``reach a triangle, then reach a circle''.
The first applies GVL on a modified environment that renders visited objects in yellow, encoding completion history directly into the observation.
The second uses VLM proposition labels to detect the sub-goal boundary and applies GVL independently on each segment.
We collect 200 optimal episodes and compute progress with each approach.
The yellow-marker environment achieves a $62\%$ relative improvement over GVL via an environment-level memory cue, while the VLM sub-goal split achieves $82\%$ by explicitly tracking sub-goal completion.
The yellow-marker fix requires a modified environment unavailable at test time; the RM's belief filter (\Cref{sec:rm-belief}) provides this memory without environment modifications.

\section{Related Work}
\label{sec:related-work}

\textbf{VLM/LLM-based reward specification.}
VLMs can judge task success zero-shot~\citep{du2023vision}, which has motivated their use as a source of reward for RL.
Early work defined rewards from the similarity between vision-language embeddings of the observation and the goal~\citep{baumli2023vision,rocamonde2023vision,sontakke2023roboclip}, whereas later work defines them from pairwise preferences over image pairs~\citep{wang2024rl}, fine-tunes a VLM into a dense reward model~\citep{lee2026roboreward}, supplies memory through test-time adaptation~\citep{ziakas2025vita}, or tracks sub-goal completion with a Bayesian filter~\citep{zhao2025t2vlm}.
GVL~\citep{ma2024vision} uses a VLM as an in-context value learner that provides dense per-timestep rewards.
\citet{venkataraman2025real} label a dataset for offline RL with a reward model trained on VLM preferences, whereas we derive rewards from reward machines whose propositions are labelled zero-shot by a pretrained VLM.
Rather than using VLM outputs directly, a complementary approach generates reward code with LLMs, as dense reward functions over the environment state~\citep{ma2024eureka,xie2024text2reward}, as reward machines with executable labelling functions~\citep{castanyer2025arm}, or as sub-goals with executable reward functions~\citep{venuto2024code}.
Alternatively, \citet{wang2024grounding,hung2024victor} combine LLM-based task decomposition with learned grounding modules.
RM-VLM uses an LLM to decompose a task into sub-task reward machines, a VLM to ground their atomic propositions, and RM states to provide memory for reward prediction.

\textbf{Formal languages for RL.}
Formal languages such as LTL~\citep{pnueli1977temporal}, whose specifications can be compiled into reward machines~\citep{camacho2019ltl}, have been widely adopted in RL for specifying temporally extended, non-Markovian reward functions.
A growing body of work trains policies that generalise zero-shot to unseen temporal logic formulas~\citep{kuo2020encoding,leon2021nutshell,vaezipoor2021ltl2action,qiu2023instructing,liu2024skill,yalcinkaya2024compositional,jackermeier2025deepltl,cloete2026platoltl,abate2026semantically,furelosblanco2026beyond}, whereas our approach requires offline RL for each new task but does not assume a ground-truth labelling function.
The methods above assume a perfect labelling function; a separate line of work relaxes this assumption.
Ground-Compose-Reinforce~\citep{li2025ground} assumes a hand-designed set of propositions and learns the labelling function from manually labelled trajectories, treating its predictions as correct when tracking the RM state.
We instead ground LLM-derived propositions with a pre-trained VLM and track a belief over RM states to absorb the resulting label noise.
Like RM-VLM, \citet{yang2024multimodal} construct an automaton with an LLM and evaluate its propositions from images with an open-vocabulary detector, but use the automaton directly as a controller instead of learning a policy.
\citet{li2024reward} and \citet{parac2024learning} mitigate the problem of noisy labelling functions by maintaining a belief over RM states, a formulation we directly build upon, but either consider only a simplified $\rmrewardtransitionfun$ structure \citep{parac2024learning} or do not specify how reward is assigned under an uncertain RM state \citep{li2024reward}, while \citet{umili2024neural} handle noise by representing the RM as a probabilistic finite automaton.
Concurrently, \citet{pannacci2026grounding} extend the approach of \citet{umili2024neural} to arbitrary LTL tasks, achieving zero-shot task composition on simple 2D environments.
Like \citet{sun2023less}, we combine reward machines with offline RL, but where they filter transitions to curate a cleaner dataset, we train on the full dataset with noisy VLM labels.
\citet{gu2026storl} shape rewards for offline RL with an LLM-generated subgoal order, but the LLM directly provides a mapping from low-dimensional states to subgoals, without using an automaton or accounting for errors in this mapping.

\textbf{From natural language to formal specifications.}
LLMs can translate natural-language instructions into temporal logic~\citep{cosler2023nl2spec,chen2023nl2tl,liu2023lang2ltl,bayat2026end} and reward machines~\citep{alsadat2025using,li2025ground,castanyer2025arm}.
These methods translate one specification at a time, over propositions given in advance or, for \citet{castanyer2025arm}, chosen per task.
We instead derive the propositions jointly from all task descriptions, so that they are shared across tasks.

\section{Conclusion}
\label{sec:conclusion}
We have presented \themethodsparse{}, a method for solving compositional tasks, including non-Markovian tasks, without a manually engineered labelling function. It uses an LLM to translate a natural-language task into an RM, a VLM to ground its atomic propositions in image observations, and a belief over RM states to mitigate labelling errors, all without fine-tuning the VLM. Across three different domains, \themethodsparse{} is shown to outperform VLM-based rewards that judge the full task, demonstrating that composing grounded propositions with an RM is more effective than querying the VLM about the task directly.

\paragraph{Limitations.}
VLM labelling accuracy varies significantly across propositions, and the method works best when propositions correspond to visually unambiguous states.
The RM state belief requires the VLM's per-proposition true- and false-positive rates, which we estimate from a small set of labelled trajectories.
Finally, RM expressiveness is limited to regular-language tasks and cannot capture properties such as counting.

\paragraph{Future work.}
In future work, we aim to extend the RM-VLM  approach to hierarchical reward machines~\citep{FurelosBlancoLJBR23}, which are effective in solving longer-horizon tasks. We plan also to learn a single policy that generalises across task specifications, considering ideas from instruction following with temporal logic~\citep{vaezipoor2021ltl2action,jackermeier2025deepltl}.

\bibliographystyle{ACM-Reference-Format}
\bibliography{references}

\clearpage
\appendix
\crefalias{section}{appendix}
\crefalias{subsection}{subappendix}

\section{Environment Descriptions}
\label{app:env_descriptions}

\subsection{GeoGrid}

\begin{figure}[h]
  \centering
  \includegraphics[width=0.4\linewidth]{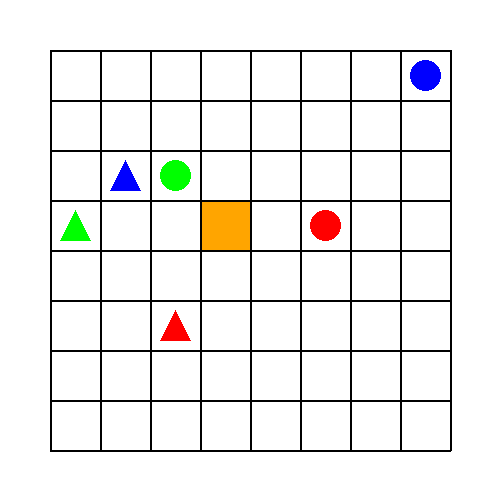}
  \caption{GeoGrid with one of the fixed layouts, as shown to the VLM. The orange cell is the agent.}
  \label{fig:geogrid_env}
\end{figure}

GeoGrid is an $8 \times 8$ grid world with six objects, a circle and a triangle in each of red, green and blue.
At the start of each episode, the objects are placed according to one of $10$ fixed layouts and the agent starts in a random empty cell.
The agent moves in the four cardinal directions.
\Cref{fig:geogrid_env} shows one of the layouts.

\paragraph{Propositions.}
The ground-truth set of propositions is given below.
Several propositions can hold at once, e.g.\ \texttt{r} and \texttt{c} when the agent stands on the red circle.

\begin{center}
\begin{tabular}{cp{0.7\columnwidth}}
\toprule
\textbf{Proposition} & \textbf{Meaning} \\
\midrule
\texttt{r} & Agent stands on a red object (any shape) \\
\texttt{g} & Agent stands on a green object (any shape) \\
\texttt{b} & Agent stands on a blue object (any shape) \\
\texttt{c} & Agent stands on a circle (any colour) \\
\texttt{t} & Agent stands on a triangle (any colour) \\
\bottomrule
\end{tabular}
\end{center}

\paragraph{Offline dataset.}
We collected $1{,}000$ episodes ($46{,}890$ transitions) of $336\times336$ images, which we use both for offline RL and for training the distilled label classifier.
Each episode follows one of the task RMs, chosen uniformly at random.
With probability $0.7$, the behaviour policy takes a step along the shortest obstacle-free path to the closest object that advances the RM, and otherwise it acts randomly.

\paragraph{VLM labelling.}
We label every frame (stride $1$) with Gemma-4-31B-it, asking about one proposition at a time with the prompt in \Cref{lst:prop-label}.
Each query includes two reference frames, one where the proposition holds and one where it does not.
Every prompt is prefixed with the same scene description (\Cref{lst:geogrid-scene}), and \Cref{lst:geogrid-questions} lists the proposition questions.

\begin{lstlisting}[style=promptstyle,
                   caption={GeoGrid scene description},
                   label={lst:geogrid-scene}]
Scene: An 8x8 grid world rendered as a top-down image.
- The grid has black grid lines on a white background.
- Exactly one cell is highlighted solid ORANGE -- this is the agent's position.
- Some cells contain objects: a coloured shape drawn centred in the cell.
- Shapes are CIRCLE or TRIANGLE.
- Colours are RED, GREEN, or BLUE.
- There are up to six objects per layout -- one for each (colour, shape) pair.
- Cells that contain neither the agent nor an object are plain white.
The agent is 'on' an object when the orange highlight occupies the same cell as that object (the coloured shape is drawn on top of the orange background).
\end{lstlisting}

\begin{lstlisting}[style=promptstyle,
                   caption={GeoGrid proposition questions (\texttt{g} and \texttt{b} are analogous to \texttt{r}, and \texttt{t} to \texttt{c})},
                   label={lst:geogrid-questions}]
r: Is the orange agent cell currently on a RED object (i.e. a red circle or red triangle drawn on top of the orange cell)?
c: Is the orange agent cell currently on a CIRCLE (of any colour -- red, green, or blue)?
\end{lstlisting}

We distil these labels into a CNN that predicts the propositions from the frame resized to $84\times84$, which is the classifier used at evaluation.
\Cref{tab:label_accuracy_geogrid} reports the accuracy of both the VLM and the distilled labels.

\begin{table}[t]
\centering
\small
\setlength{\tabcolsep}{4pt}
\caption{%
  Per-proposition label accuracy (\%) on GeoGrid.
  \textbf{VLM}: Gemma-4-31B-it labels.
  \textbf{Distilled}: the CNN distilled from them, used at evaluation.
  \% Positive is the proportion of frames where each proposition holds, and Overall is micro-averaged over propositions.
}
\label{tab:label_accuracy_geogrid}
\begin{tabular}{lcccccc}
\toprule
 & \textbf{r} & \textbf{g} & \textbf{b} & \textbf{c} & \textbf{t} & \textbf{Overall} \\
\midrule
\% Positive & 5.5 & 5.1 & 5.1 & 4.3 & 11.4 & 6.3 \\
\midrule
\multicolumn{7}{l}{\emph{VLM}} \\
Accuracy  & 99.6 & 99.7 & 99.7 & 99.4 & 99.5 & 99.6 \\
Precision & 94.0 & 95.1 & 94.7 & 88.5 & 96.7 & 94.5 \\
Recall    & 98.9 & 98.7 & 99.1 & 98.7 & 98.8 & 98.9 \\
\midrule
\multicolumn{7}{l}{\emph{Distilled}} \\
Accuracy  & 100.0 & 100.0 & 100.0 & 100.0 & 100.0 & 100.0 \\
Precision & 100.0 & 99.8 & 100.0 & 100.0 & 100.0 & 100.0 \\
Recall    & 100.0 & 100.0 & 100.0 & 100.0 & 100.0 & 100.0 \\
\bottomrule
\end{tabular}
\end{table}

\subsection{OGBench-Scene}

\begin{figure}[h]
  \centering
  \includegraphics[width=0.5\linewidth]{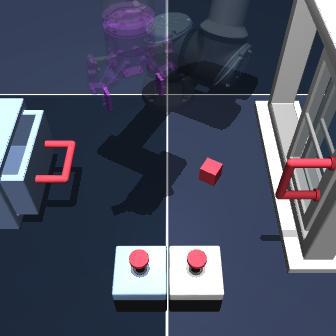}
  \caption{OGBench-Scene as shown to the VLM ($336\times336$ frame, robot arm drawn transparent). Red handles and buttons indicate that the drawer and window are locked.}
  \label{fig:scene_env}
\end{figure}

OGBench-Scene~\citep{park2024ogbench} is a robotic manipulation environment.
It contains a \emph{drawer}, a \emph{window}, two \emph{lock buttons} and a solid red cube.
Its visualisation can be seen in \Cref{fig:scene_env}.

\paragraph{Propositions.}
The ground-truth set of propositions is given below.

\begin{center}
\begin{tabular}{cp{0.7\columnwidth}}
\toprule
\textbf{Proposition} & \textbf{Meaning} \\
\midrule
\texttt{du} & Drawer is unlocked (its handle is not red; toggled by the left button) \\
\texttt{wu} & Window is unlocked (its handle is not red; toggled by the right button) \\
\texttt{do} & Drawer is open \\
\texttt{wo} & Window is fully open \\
\texttt{ci} & Cube is inside the drawer \\
\bottomrule
\end{tabular}
\end{center}

\paragraph{Offline dataset.}
We use OGBench's \texttt{visual-scene-play} dataset ($1{,}000$ trajectories, $10^6$ transitions).
We re-rendered its images at $336\times336$ from the stored simulator states.

\paragraph{VLM labelling.}
We label the frames with Gemma-4-31B-it in the same way as in GeoGrid, except that the VLM sees every $10$th frame and we interpolate the labels in between.
Every prompt is prefixed with the same scene description (\Cref{lst:scene-description}).

\begin{lstlisting}[style=promptstyle,
                   caption={OGBench-Scene scene description},
                   label={lst:scene-description}]
Scene layout:
- LEFT side: a drawer (white box with a sliding front panel).
- RIGHT side: a window (a sliding panel in a frame).
- BOTTOM CENTER: two lock buttons on pedestals. The LEFT button corresponds to the drawer lock; the RIGHT button corresponds to the window lock. A button is RED when the corresponding object is LOCKED, and WHITE/GREY when it is UNLOCKED.
- The window also has a handle bar on its frame that turns RED when locked.
- A solid RED cube is somewhere in the scene -- on the floor or inside the drawer. If the cube is not visible, it is inside the drawer. - When the drawer is CLOSED, the front panel and handle sit flush against the box.
- When the drawer is OPEN, the panel is displaced to one side, revealing a gap and the interior cavity.
- When the window is CLOSED, the handle bar sits at the top of the window frame, far from the buttons.
- When the window is OPEN, the handle has slid down so the handle bar is near the buttons.
- A transparent robot arm with a purple gripper which performs the manipulation.
\end{lstlisting}

\Cref{lst:scene-questions} lists the proposition questions.

\begin{lstlisting}[style=promptstyle,
                   caption={OGBench-Scene proposition questions (\texttt{wu} is analogous to \texttt{du})},
                   label={lst:scene-questions}]
du: Is the drawer unlocked? (The drawer is locked if its handle is red; it is unlocked if the handle is not red. The drawer gets unlocked/locked by pressing the left button on the table.)
do: Is the drawer open?
wo: Is the window fully open? (A partially open window does not count -- the window must be fully open with the handle slid all the way down near the buttons.)
ci: Is the cube inside the drawer?
\end{lstlisting}

We distil these labels into a CNN that predicts the propositions from the frame resized to $64\times64$, which is the classifier used at evaluation.
\Cref{tab:label_accuracy_scene_gemma4_31b_train} reports the accuracy of both the VLM and the distilled labels.

\begin{table}[t]
\centering
\small
\setlength{\tabcolsep}{4pt}
\caption{%
  Per-proposition label accuracy (\%) on OGBench-Scene.
  \textbf{VLM}: Gemma-4-31B-it labels.
  \textbf{Distilled}: the CNN distilled from them, used at evaluation.
  \% Positive is the proportion of frames where each proposition holds, and Overall is micro-averaged over propositions.
}
\label{tab:label_accuracy_scene_gemma4_31b_train}
\begin{tabular}{lcccccc}
\toprule
 & \textbf{du} & \textbf{wu} & \textbf{do} & \textbf{wo} & \textbf{ci} & \textbf{Overall} \\
\midrule
\% Positive & 51.8 & 54.5 & 31.9 & 36.1 & 17.0 & 38.3 \\
\midrule
\multicolumn{7}{l}{\emph{VLM}} \\
Accuracy  & 98.3 & 97.0 & 74.6 & 59.0 & 99.3 & 85.7 \\
Precision & 98.7 & 98.4 & 56.3 & 42.8 & 97.8 & 78.9 \\
Recall    & 98.0 & 96.2 & 90.1 & 40.3 & 98.4 & 85.3 \\
\midrule
\multicolumn{7}{l}{\emph{Distilled}} \\
Accuracy  & 98.9 & 98.3 & 81.7 & 58.3 & 99.2 & 87.3 \\
Precision & 99.5 & 99.1 & 64.2 & 41.4 & 97.2 & 81.5 \\
Recall    & 98.2 & 97.7 & 96.5 & 37.6 & 98.0 & 86.3 \\
\bottomrule
\end{tabular}
\end{table}

\subsection{MiniWorld-Pickup}

\begin{figure}[h]
  \centering
  \includegraphics[width=0.5\linewidth]{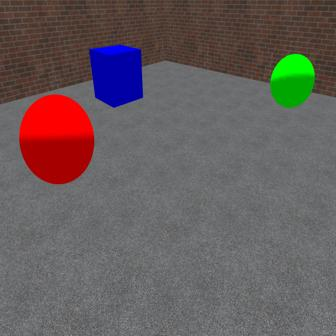}
  \caption{MiniWorld-Pickup from the agent's egocentric view, as shown to the VLM ($336\times336$ frame).}
  \label{fig:miniworld_env}
\end{figure}

MiniWorld~\citep{chevalier2023minigrid} is a library of configurable, egocentric and partially observable 3D environments.
MiniWorld-Pickup, which we build with it, contains a walled room with six objects (red box, red ball, blue box, blue ball, green box, green ball), placed according to one of $10$ fixed layouts.
The agent can turn left or right, move forward or backward, or pick up an object in front of it.
The objects are consumed on pickup and no longer appear in the environment.
\Cref{fig:miniworld_env} shows the agent's view.

\paragraph{Propositions.}
Unlike in the other two environments, the propositions here are \emph{pickup events} rather than properties of the state.
A proposition is true only on the step where the agent picks up a matching object.
Several can be true at once, e.g.\ picking up a red box makes both \texttt{r} and \texttt{x} true.

\begin{center}
\begin{tabular}{cp{0.7\columnwidth}}
\toprule
\textbf{Proposition} & \textbf{Meaning} \\
\midrule
\texttt{r} & A red object (box or ball) was just picked up \\
\texttt{g} & A green object (box or ball) was just picked up \\
\texttt{b} & A blue object (box or ball) was just picked up \\
\texttt{x} & A box (any colour) was just picked up \\
\texttt{o} & A ball (any colour) was just picked up \\
\bottomrule
\end{tabular}
\end{center}

\paragraph{Offline dataset.}
For offline RL and for training the distilled event detector, we collected a multi-task dataset with $5{,}000$ episodes ($728{,}355$ transitions) where the state is a $336\times336$ egocentric image, with a horizon of $800$ steps.
Each episode follows one of the task RMs, with the sequence, safety, logic and alternation tasks drawn in a $1{:}2{:}2{:}2$ ratio.
With probability $0.9$, the behaviour policy walks to the closest object that advances the RM without rejecting and picks it up, and otherwise it acts randomly.

\paragraph{VLM labelling.}
We label the episodes with Gemma-4-31B-it.
Eight episodes do not fit in its $262{,}144$-token context, so we leave them unlabelled and keep $4{,}992$ of the $5{,}000$.
Every prompt is prefixed with the same scene description:

\begin{lstlisting}[style=promptstyle,
                   caption={MiniWorld-Pickup scene description},
                   label={lst:miniworld-scene}]
Scene: A first-person (egocentric) view inside a single 3D room with brick walls and an asphalt floor. Scattered on the floor are coloured objects: BOXES (cubes) and BALLS (spheres), in RED, GREEN, and BLUE. The camera is tilted slightly downward. An object is 'close and centered in front' when it is large and near the bottom-centre of the view (the agent is right up against it). Objects that have been picked up are removed from the room.
\end{lstlisting}

Since the propositions are pickup events, we ask the VLM \emph{when each object disappears}.
As in the other environments, we query each object separately, giving six queries per episode.
Each query shows every frame of the episode (stride $1$), numbered by its original index, followed by the disappearance question for that object (\Cref{lst:miniworld-disappear} shows it for the red box).

\begin{lstlisting}[style=promptstyle,
                   caption={MiniWorld-Pickup object disappearance question (red box)},
                   label={lst:miniworld-disappear}]
At which frame does the red box disappear from the environment? The red box can disappear AT MOST ONCE in the episode: the agent consumes it by walking right up to it and picking it up. The disappearance looks like this: in one frame the red box is large and centred directly in front of the agent, and in the next frame it is gone while the camera has NOT moved. IMPORTANT: the red box merely being out of view (because the agent looked or walked away) is NOT disappearance -- only count it if it vanishes while the view stays still. Conversely, if the agent walks right up to the red box until it is large in the centre of the view, and the object is then never visible again in ANY later frame, that IS a consumption -- do not miss these. Answer with the frame index of the FIRST frame in which the red box has vanished, or -1 if it never disappears.
\end{lstlisting}

Each reported disappearance frame is snapped to the nearest \texttt{pickup} action within $\pm 4$ frames to pin the exact transition, yielding per-proposition pickup events on $\{$\texttt{r},\texttt{g},\texttt{b},\texttt{x},\texttt{o}$\}$.

We then distil these events into a small CNN that detects pickups from two consecutive $64\times64$ frames.
We apply this detector only at frames where the agent takes the pickup action.
The detector labels the evaluation episodes, and it also provides the labels for offline RL.
To track the RM state we use the belief, setting each proposition's noise to the detector's measured true- and false-positive rates.
\Cref{tab:label_accuracy_miniworld} reports the accuracy of both the VLM and the distilled labels.

\begin{table}[t]
\centering
\small
\setlength{\tabcolsep}{4pt}
\caption{%
  Per-proposition label accuracy (\%) on MiniWorld-Pickup.
  \textbf{VLM}: Gemma-4-31B-it labels.
  \textbf{Distilled}: the pickup detector distilled from them, used at evaluation.
  \% Positive is the proportion of frames where each proposition holds, and Overall is micro-averaged over propositions.
  Positives are rare, so precision and recall are more informative than accuracy.
}
\label{tab:label_accuracy_miniworld}
\begin{tabular}{lcccccc}
\toprule
 & \textbf{r} & \textbf{g} & \textbf{b} & \textbf{x} & \textbf{o} & \textbf{Overall} \\
\midrule
\% Positive & 0.9 & 0.8 & 0.8 & 1.1 & 1.4 & 1.0 \\
\midrule
\multicolumn{7}{l}{\emph{VLM}} \\
Accuracy  & 100.0 & 100.0 & 100.0 & 100.0 & 99.9 & 100.0 \\
Precision & 99.2 & 99.6 & 99.7 & 99.1 & 98.3 & 99.0 \\
Recall    & 98.3 & 96.8 & 95.8 & 96.4 & 97.4 & 97.0 \\
\midrule
\multicolumn{7}{l}{\emph{Distilled}} \\
Accuracy  & 100.0 & 100.0 & 100.0 & 100.0 & 100.0 & 100.0 \\
Precision & 99.6 & 99.2 & 99.7 & 98.8 & 98.7 & 99.1 \\
Recall    & 99.0 & 98.9 & 98.8 & 98.4 & 98.9 & 98.8 \\
\bottomrule
\end{tabular}
\end{table}

\section{Proposition, Reward-Machine and Label Prior Generation}
\label{app:rm-construction}

We evaluate the generation steps of \Cref{sec:grounded-rms,sec:rm-belief} with Gemma-4-31B-it on the tasks in \Cref{tab:tasks}, over 10 runs per task, using the recommended sampling settings (temperature $1.0$, top-$p$ $0.95$, top-$k$ $64$)~\citep{team2026gemma4}.
The prompts for the three steps are given in \Cref{lst:prop-gen,lst:rm-gen,lst:prior-gen}.
Every prompt includes the environment's scene description and possibly its physical constraints (\Cref{lst:env-constraints}).

\paragraph{Proposition generation.}
The LLM generates one set of propositions from all task descriptions of an environment (\Cref{lst:prop-gen}).
Generated propositions can differ from the reference ones in naming or factorisation, e.g.\ on GeoGrid the LLM generates six object propositions instead of three colours and two shapes.
To compare them, the same LLM at temperature $0$ writes each generated proposition as a Boolean formula over the reference propositions.
Across runs and environments, the generated propositions can express all reference propositions the tasks require.
In addition, all but one of the $173$ generated propositions appear in at least one of the RMs later generated with them.

\paragraph{RM generation.}
We additionally use the LLM to generate the RM for each task given the propositions (\Cref{lst:rm-gen}).
We test two prompting strategies: \emph{decomposed}, where the LLM produces one RM per sub-task, and \emph{monolithic}, where the LLM produces a single RM for the full task.
Each generated RM is compared with the expert-written reference RM.
Two RMs are equivalent if they have the same set of minimal accepting label sequences up to length $L=15$, and for the cyclic Loop task we compare the sequences that complete one cycle.
For this check, decomposed RMs are composed via cross-product~\citep{neary2020reward}, and for generated propositions, the reference label sequences are translated into the generated alphabet using the formulas above.
Given the reference propositions, the decomposed strategy produces an equivalent RM in every run, while the monolithic strategy fails on OGBench-Scene tasks that have multiple valid orderings as well as occasionally on MiniWorld-Pickup (\Cref{tab:rm_construction}).
Over the generated propositions, the decomposed strategy does so in $114/120$ runs, and the failures are malformed RMs, such as non-deterministic ones.

\paragraph{Label prior generation.}
For each reference RM, the LLM generates the label prior $P(\labelobs \mid \rmstate)$ of the RM state belief (\Cref{lst:prior-gen}), from the same environment description and physical constraints as RM generation.
It first classifies each proposition as \emph{lasting} or \emph{momentary}.
A lasting proposition holds until something changes it, and a momentary one holds only at the step its event happens.
The LLM also gives each proposition's value at the start of the episode.
For each RM state $\rmstate$ and lasting proposition $\proposition$, it then answers two questions: whether $\proposition$ is true when the RM reaches $\rmstate$, and whether a single action can change $\proposition$ at the next step.
A momentary proposition is false and can change in every state.
We map each pair of answers to the probability $P(\labelobs_\proposition{=}1 \mid \rmstate)$ that $\proposition$ is true in $\rmstate$, namely $1$ (true, cannot change), $1-\epsilon$ (true, can change), $\epsilon$ (false, can change) or $0$ (false, cannot change), so a change that triggers an RM transition never has zero probability.
Finally, the LLM lists the combinations of momentary propositions that can hold at a single step.
The prior $P(\labelobs \mid \rmstate)$ is proportional to $\prod_{\proposition} P(\labelobs_\proposition \mid \rmstate)$ over the label sets whose momentary part is listed, and $0$ otherwise.
The $\epsilon$ value is considered a hyperparameter, but by setting it to $\epsilon = 0.005$ on OGBench-Scene and $\epsilon = 0.116$ on GeoGrid and MiniWorld-Pickup, we can recover the expert-written reference.
In fact, the LLM-generated prior matches the expert-written one in $119/120$ runs, with the only error coming from a GeoGrid Logic run whose response does not follow the required format.

\begin{table}[t]
\centering
\small
\setlength{\tabcolsep}{2.3pt}
\caption{%
  RM generation accuracy: fraction of runs producing a trace-equivalent RM (10 runs per cell, Gemma-4-31B-it). \emph{Decomposed} and \emph{Monolithic} use the reference propositions; \emph{Gen.\ props} generates the propositions first and uses the decomposed prompt.}
\label{tab:rm_construction}
\begin{tabular}{llccc}
\toprule
\textbf{Env} & \textbf{Task} & \textbf{Decomposed} & \textbf{Monolithic} & \textbf{Gen.\ props} \\
\midrule
\multirow{4}{*}{GeoGrid}
 & Sequence & $10/10$ & $10/10$ & $10/10$ \\
 & Loop     & $10/10$ & $10/10$ & $10/10$ \\
 & Logic    & $10/10$ & $10/10$ & $10/10$ \\
 & Safety   & $10/10$ & $10/10$ & $10/10$ \\
\midrule
\multirow{4}{*}{\shortstack[l]{OGBench-\\Scene}}
 & Open           & $10/10$ & $0/10$  & $8/10$ \\
 & Close+Lock     & $10/10$ & $0/10$  & $10/10$ \\
 & Cube-in-Drawer & $10/10$ & $10/10$ & $10/10$ \\
 & Cube+Window    & $10/10$ & $0/10$  & $10/10$ \\
\midrule
\multirow{4}{*}{\shortstack[l]{MiniWorld-\\Pickup}}
 & Sequence    & $10/10$ & $10/10$ & $10/10$ \\
 & Logic       & $10/10$ & $8/10$  & $7/10$ \\
 & Safety      & $10/10$ & $9/10$  & $10/10$ \\
 & Alternation & $10/10$ & $10/10$ & $9/10$ \\
\bottomrule
\end{tabular}
\end{table}

\section{Reduction of Belief Reward to RM Reward}
\label{app:one-hot-proof}

\begin{prop}
\label{prop:one-hot}
If the RM state belief $\rmbelief_t$ is one-hot for all $t$ (i.e.\ $\rmbelief_t[\rmstate^*_t] = 1$ for some $\rmstate^*_t \in \rmstates$), then the task reward (\Cref{eq:task-reward}) reduces to the standard RM transition reward:
\begin{equation*}
\rmrewardtransitionfun(\rmbelief_{t-1}, \rmbelief_t) = \rmrewardtransitionfun(\rmstate^*_{t-1},\, \rmstate^*_t).
\end{equation*}
\end{prop}

\begin{proof}
Since $\rmbelief_{t-1}$ is one-hot at $\rmstate^*_{t-1}$, the numerator and denominator of \Cref{eq:alpha-update} retain only the $\rmstate = \rmstate^*_{t-1}$ term.
Evaluating at $\rmstate' = \rmstate^*_t$:
\begin{align*}
\rmalpha_t[\rmstate^*_t]
  &= \frac{C[\rmstate^*_{t-1}, \rmstate^*_t, \noisylabelobs_{t-1}]\,
       \bigl(\rmalpha_{t-1}[\rmstate^*_{t-1}] + \rmrewardtransitionfun(\rmstate^*_{t-1}, \rmstate^*_t)\bigr)}
      {C[\rmstate^*_{t-1}, \rmstate^*_t, \noisylabelobs_{t-1}]} \\
  &= \rmalpha_{t-1}[\rmstate^*_{t-1}] + \rmrewardtransitionfun(\rmstate^*_{t-1}, \rmstate^*_t).
\end{align*}
Substituting into \Cref{eq:task-reward} with one-hot beliefs:
\begin{align*}
\rmrewardtransitionfun(\rmbelief_{t-1}, \rmbelief_t)
  &= (\rmbelief_t)^\top \rmalpha_t - (\rmbelief_{t-1})^\top \rmalpha_{t-1} \\
  &= \rmalpha_t[\rmstate^*_t] - \rmalpha_{t-1}[\rmstate^*_{t-1}] \\
  &= \rmrewardtransitionfun(\rmstate^*_{t-1}, \rmstate^*_t). \qedhere
\end{align*}
\end{proof}

\section{RM State Belief: Worked Example}
\label{app:belief-grid-example}

\begin{example}[Correlated propositions absorb VLM errors]
\label{ex:belief-grid}
Consider a ``reach red'' RM on GeoGrid (\Cref{app:env_descriptions}) with the full proposition set $\propositions = \set{\propr,\propg,\propb,\propc,\propt}$.
The prior $P(\labelobs \mid \initialrmstate)$ encodes that each object has both a colour and a shape:
\begin{gather*}
P(\emptyset \mid \initialrmstate) = \tfrac{58}{64}, \\
P(\set{\propr,\propc} \mid \initialrmstate) = P(\set{\propr,\propt} \mid \initialrmstate) = P(\set{\propg,\propc} \mid \initialrmstate) \\
= P(\set{\propg,\propt} \mid \initialrmstate) = P(\set{\propb,\propc} \mid \initialrmstate) = P(\set{\propb,\propt} \mid \initialrmstate) = \tfrac{1}{64},
\end{gather*}
with $P(\labelobs \mid \initialrmstate) = 0$ for all other $\labelobs$. In particular, $\propr$ alone is structurally impossible.
Suppose that the per-proposition classifier rates are $\mathrm{TPR}_\proposition = 0.9$ and $\mathrm{FPR}_\proposition = 0.05$ for all $\proposition$, and the VLM predicts $\noisylabelobs_0 = \set{\propr}$.
In this case, the VLM incorrectly predicts $\propr$ as true but correctly predicts all other propositions as false.
If deterministic tracking were used, this false positive would immediately trigger a transition to the accepting state.
With the RM state belief update, we compute $\hat{\rmbelief}_1$ via \Cref{eq:belief-update}:
\begin{align*}
C[\initialrmstate, \initialrmstate, \set{\propr}]
  &= \underbrace{\tfrac{58}{64} \cdot 0.05 \cdot 0.95^4}_{\labelobs = \emptyset} \\
  &\quad + \underbrace{\tfrac{1}{16} \cdot 0.05 \cdot 0.1^2 \cdot 0.95^2}_{%
    \substack{\labelobs \in \{\{\propg,\propc\},\{\propg,\propt\},\\\{\propb,\propc\},\{\propb,\propt\}\}}}
  \approx 0.037, \nonumber \\[6pt]
C[\initialrmstate, \acceptingrmstate, \set{\propr}]
  &= \underbrace{\tfrac{1}{64} \cdot 0.9 \cdot 0.1 \cdot 0.95^3}_{\labelobs = \set{\propr,\propc}} \\
  &\quad + \underbrace{\tfrac{1}{64} \cdot 0.9 \cdot 0.1 \cdot 0.95^3}_{\labelobs = \set{\propr,\propt}}
  \approx 0.002. \nonumber
\end{align*}

After normalisation, $\rmbelief_1[\initialrmstate] \approx 0.94$.
Despite the erroneous prediction, the belief is still mostly concentrated on the initial state.
In contrast, if the VLM predicts $\noisylabelobs_0 = \set{\propr,\propc}$, which is consistent with the prior, the balance shifts:
\begin{align*}
C[\initialrmstate, \initialrmstate, \set{\propr,\propc}]
  &= \underbrace{\tfrac{58}{64} \cdot 0.05^2 \cdot 0.95^3}_{\labelobs = \emptyset} \\
  &\quad + \underbrace{2 \cdot \tfrac{1}{64} \cdot 0.05 \cdot 0.1 \cdot 0.9 \cdot 0.95^2}_{\labelobs \in \set{\set{\propg,\propc},\set{\propb,\propc}}} \\
  &\quad + \underbrace{2 \cdot \tfrac{1}{64} \cdot 0.05^2 \cdot 0.1^2 \cdot 0.95}_{\labelobs \in \set{\set{\propg,\propt},\set{\propb,\propt}}}
  \approx 0.002, \\[6pt]
C[\initialrmstate, \acceptingrmstate, \set{\propr,\propc}]
  &= \underbrace{\tfrac{1}{64} \cdot 0.9^2 \cdot 0.95^3}_{\labelobs = \set{\propr,\propc}} \\
  &\quad + \underbrace{\tfrac{1}{64} \cdot 0.9 \cdot 0.05 \cdot 0.1 \cdot 0.95^2}_{\labelobs = \set{\propr,\propt}} \\
  &\qquad \approx 0.011. \nonumber
\end{align*}
After normalisation, $\rmbelief_1[\acceptingrmstate] \approx 0.84$. As the observation is consistent with the prior, the belief shifts towards the accepting state.
\end{example}

\section{Choice of Offline RL Algorithm}
\label{app:scene-fql}
\label{sec:algorithm-generalisation}

To test whether the benefits of \themethodsparse{} are tied to IQL, we repeat the OGBench-Scene evaluation with FQL (\Cref{tab:scene_fql}).
\themethodsparse{} again outperforms the non-RM baselines on all OGBench-Scene tasks, showing that the gains are not specific to IQL.
Adding dense VLM progress within transitions (\themethoddense) performs on par with or slightly above \themethodsparse{} alone on three of four tasks and slightly below it on Cube-in-Drawer.
FQL solves Close+Lock ($0.92$) from the same VLM-based reward on which IQL reaches $0.53$, indicating that it is more robust to the labelling errors that limit IQL on this task (\Cref{tab:ablation_belief}).
Conversely, FQL performs worse than IQL on the two cube tasks.

\begin{table}[t]
\centering
\small
\setlength{\tabcolsep}{3pt}
\caption{OGBench-Scene task success rates with FQL~\citep{park2025fql} (mean $\pm$ std over 5 seeds). Values within $95\%$ of the best per row in bold.}
\label{tab:scene_fql}
\resizebox{\columnwidth}{!}{%
\begin{tabular}{lccccc}
\toprule
\textbf{Task} & \textbf{BC} & \textbf{GVL} & \textbf{VLM-SD} & \textbf{\themethodsparse} & \textbf{\themethoddense} \\
\midrule
Open & $0.60 \pm 0.05$ & $0.73 \pm 0.12$ & $0.32 \pm 0.12$ & $\mathbf{0.99 \pm 0.01}$ & $\mathbf{0.97 \pm 0.02}$ \\
Close+Lock & $0.09 \pm 0.03$ & $0.13 \pm 0.08$ & $0.10 \pm 0.04$ & $\mathbf{0.92 \pm 0.04}$ & $\mathbf{0.93 \pm 0.06}$ \\
Cube-in-Drawer & $0.04 \pm 0.02$ & $0.04 \pm 0.04$ & $0.02 \pm 0.02$ & $\mathbf{0.61 \pm 0.06}$ & $0.56 \pm 0.08$ \\
Cube+Window & $0.01 \pm 0.01$ & $0.02 \pm 0.01$ & $0.03 \pm 0.02$ & $0.47 \pm 0.07$ & $\mathbf{0.51 \pm 0.09}$ \\
\bottomrule
\end{tabular}}
\end{table}

\begin{table*}[t]
\centering
\caption{Offline RL (IQL) hyperparameters for the main results. All methods within an environment share these values, except that the GeoGrid and MiniWorld-Pickup baselines use a CNN-GRU encoder.}
\label{tab:hyperparams}
\begin{tabular}{lccc}
\toprule
\textbf{Hyperparameter} & \textbf{GeoGrid} & \textbf{OGBench-Scene} & \textbf{MiniWorld-Pickup} \\
\midrule
Training steps               & $10^6$              & $10^6$ & $7\times10^5$ \\
Batch size                   & $256$               & $256$ & $256$ \\
Optimiser                    & Adam                & Adam & Adam \\
Learning rate                & $3\times10^{-4}$    & $3\times10^{-4}$ & $3\times10^{-4}$ \\
Discount $\discountfactor$   & $0.99$              & $0.99$ & $0.99$ \\
IQL expectile                & $0.7$               & $0.85$ & $0.6$ \\
Advantage-weighting temperature & $3.0$            & $10.0$ & $3.0$ \\
Max advantage weight         & $100$               & $100$ & $100$ \\
Target-network EMA rate      & $0.005$             & $0.005$ & $0.005$ \\
Observation / encoder        & $84{\times}84$ image, CNN & $64{\times}64$ image (3-frame stack), CNN & $64{\times}64$ image (3-frame stack), CNN \\
Baseline encoder             & CNN-GRU ($512$)     & CNN (as above) & CNN-GRU ($512$) \\
Reward scaler                & return-based ($\times 10^3$) & none & return-based ($\times 10^3$) \\
RM value-iteration discount $\rmdiscount$ & $0.9$   & $0.9$ & $0.9$ \\
\bottomrule
\end{tabular}
\end{table*}

\section{Training Details and Hyperparameters}
\label{app:hyperparams}

We list the hyperparameters used for offline RL with IQL~\citep{kostrikov2021offline} in \Cref{tab:hyperparams}.
\Cref{tab:hyperparams_other} lists the hyperparameters of the FQL results on OGBench-Scene (\Cref{tab:scene_fql}) and of the online PPO results on GeoGrid (\Cref{sec:online-rl}).

\paragraph{Evaluation.}
For each run, we select the checkpoint with the best training-time evaluation and report its success on $200$ held-out episodes whose seeds are disjoint from those used for selection.
Training-time evaluation runs every $5 \times 10^4$ steps on GeoGrid ($10$ episodes, one per layout), every $2 \times 10^4$ steps on OGBench-Scene ($50$ episodes) and every $10^5$ steps on MiniWorld-Pickup ($50$ episodes).

When doing the RM state belief ablation (\Cref{tab:ablation_belief}), the Naive and IBU variants use the un-augmented RMs on OGBench-Scene, as the extra RM states (e.g.\ \emph{unlocked}) exist only to support the state-dependent structural prior.
On GeoGrid and MiniWorld-Pickup all methods use the same RM.

\paragraph{Compute.}
We used a cluster on NVIDIA H200 GPUs.
A single offline run costs about $2.5$ GPU-hours on GeoGrid, $4$ on OGBench-Scene with IQL and $3$ with FQL, and $8.5$ on MiniWorld-Pickup; an online PPO run on GeoGrid costs about $3$.
Labelling with Gemma-4-31B-it, served by SGLang in FP8, costs about $70$ GPU-hours for the propositions and $70$ for the GVL progress estimates on OGBench-Scene, $125$ and $120$ respectively on MiniWorld-Pickup, and roughly $30$ each on GeoGrid, which was labelled on a GH200 GPU.

\begin{table}[t]
\centering
\small
\caption{Hyperparameters of the FQL results on OGBench-Scene (\Cref{tab:scene_fql}) and of the online PPO results on GeoGrid (\Cref{sec:online-rl}).}
\label{tab:hyperparams_other}
\begin{tabular}{lc}
\toprule
\textbf{Hyperparameter} & \textbf{Value} \\
\midrule
\multicolumn{2}{l}{\emph{FQL, OGBench-Scene}} \\
Training steps               & $10^6$ \\
Batch size                   & $256$ \\
Learning rate                & $3\times10^{-4}$ \\
Discount $\discountfactor$   & $0.99$ \\
Target-network EMA rate      & $0.005$ \\
BC coefficient $\alpha$      & $3$ \\
Flow steps                   & $10$ \\
Target $Q$ aggregation       & mean \\
$Q$-loss normalisation       & yes \\
Reward scale                 & $100$ \\
Actor / critic MLP           & $4 \times 512$ \\
Observation / encoder        & $64{\times}64$, 3-frame stack, CNN \\
Evaluation interval / episodes & $2\times10^4$ / $50$ \\
\midrule
\multicolumn{2}{l}{\emph{PPO, online GeoGrid}} \\
Environment steps            & $10^7$ \\
Parallel environments        & $32$ \\
Rollout length (per env.)    & $2048$ \\
Minibatch size               & $8192$ \\
Epochs per update            & $10$ \\
Learning rate                & $3\times10^{-4}$ \\
Discount $\discountfactor$   & $0.99$ \\
GAE $\lambda$                & $0.95$ \\
Clip range                   & $0.2$ \\
Entropy coefficient          & $0.05$ \\
Value-loss coefficient       & $0.5$ \\
Max gradient norm            & $0.5$ \\
Observation / encoder        & $84{\times}84$, CNN \\
RM value-iteration discount $\rmdiscount$ & $0.9$ \\
\bottomrule
\end{tabular}
\end{table}

\section{Controlled Label Noise on GeoGrid}
\label{app:label-noise}

The VLM labels on GeoGrid are nearly perfect (\Cref{tab:label_accuracy_geogrid}), so all RM tracking variants in \Cref{tab:ablation_belief} perform equally well.
We therefore inject synthetic noise into the ground-truth labels of the GeoGrid Logic and Safety tasks and train the three variants of \Cref{sec:ablations} on them.
Each proposition label in the dataset is flipped independently with probability $\epsilon \in \{0, 0.005, 0.02, 0.05\}$.
At evaluation, the labels fed to the policy's RM tracking are flipped at the same rate, while success is judged on the true RM state.
Under Naive tracking, these rates corrupt the RM state in $15\%$, $46\%$ and $75\%$ of the Logic episodes and $11\%$, $37\%$ and $64\%$ of the Safety episodes.
We follow the setting of the GeoGrid runs in \Cref{tab:main_results}, only setting the $\mathrm{TPR}_\proposition = 1 - \epsilon$ and $\mathrm{FPR}_\proposition = \epsilon$ ($0.99$ and $0.005$ at $\epsilon = 0$).
For evaluation, we report success on $200$ held-out episodes, with the best checkpoint by $20$ evaluations of $10$ episodes.

\begin{table}[t]
\centering
\small
\caption{Task success on GeoGrid under synthetic label noise (mean $\pm$ std over 5 seeds). $\epsilon$ is the probability of flipping each proposition label. Values within $95\%$ of the best per row in bold.}
\label{tab:label_noise}
\begin{tabular}{lccc}
\toprule
$\epsilon$ & \textbf{Naive} & \textbf{IBU} & \textbf{Belief} \\
\midrule
\multicolumn{4}{l}{\emph{Logic}} \\
$0$     & $\mathbf{0.98 \pm 0.02}$ & $0.91 \pm 0.05$ & $\mathbf{0.99 \pm 0.00}$ \\
$0.005$ & $0.84 \pm 0.01$ & $0.74 \pm 0.06$ & $\mathbf{0.91 \pm 0.04}$ \\
$0.02$  & $0.55 \pm 0.02$ & $0.00 \pm 0.00$ & $\mathbf{0.79 \pm 0.02}$ \\
$0.05$  & $0.16 \pm 0.02$ & $0.00 \pm 0.00$ & $\mathbf{0.52 \pm 0.02}$ \\
\midrule
\multicolumn{4}{l}{\emph{Safety}} \\
$0$     & $\mathbf{0.95 \pm 0.01}$ & $\mathbf{0.96 \pm 0.01}$ & $\mathbf{0.94 \pm 0.02}$ \\
$0.005$ & $0.82 \pm 0.02$ & $0.80 \pm 0.02$ & $\mathbf{0.90 \pm 0.02}$ \\
$0.02$  & $0.70 \pm 0.03$ & $0.75 \pm 0.02$ & $\mathbf{0.85 \pm 0.01}$ \\
$0.05$  & $0.42 \pm 0.03$ & $0.59 \pm 0.02$ & $\mathbf{0.79 \pm 0.03}$ \\
\bottomrule
\end{tabular}
\end{table}

\begin{table}[t]
\centering
\small
\caption{Belief on GeoGrid Logic with a misspecified noise model (mean $\pm$ std over 5 seeds). The assumed flip rate is $0.5\times$, $1\times$ or $2\times$ the true rate $\epsilon$.}
\label{tab:label_noise_misspec}
\begin{tabular}{lccc}
\toprule
$\epsilon$ & \textbf{Assumed $0.5\times$} & \textbf{Assumed $1\times$} & \textbf{Assumed $2\times$} \\
\midrule
$0.02$ & $\mathbf{0.79 \pm 0.02}$ & $\mathbf{0.79 \pm 0.02}$ & $\mathbf{0.77 \pm 0.02}$ \\
$0.05$ & $\mathbf{0.54 \pm 0.05}$ & $\mathbf{0.52 \pm 0.02}$ & $0.00 \pm 0.00$ \\
\bottomrule
\end{tabular}
\end{table}

\Cref{tab:label_noise} reports the results.
Belief degrades most gracefully on both tasks, and its gap to Naive widens with the noise, from $+0.07$ to $+0.36$ on Logic and from $+0.08$ to $+0.37$ on Safety.
IBU collapses on Logic from $\epsilon = 0.02$, as its flat prior spreads the belief over RM states, so the structural prior accounts for the remaining gain.
We also evaluate the impact of inaccurately specifying the TPR and FPR used by Belief, assuming half or twice the true flip rate on GeoGrid Logic (\Cref{tab:label_noise_misspec}).
Halving the rate does not reduce success, and doubling it costs only $0.02$ at $\epsilon = 0.02$.
At $\epsilon = 0.05$, however, the doubled rate makes the belief too diffuse to learn from.

\FloatBarrier
\section{Prompts}
\label{app:prompts}

This appendix collects the prompts used throughout our pipeline.

\subsection{Proposition generation}

\begin{lstlisting}[style=promptstyle,
                   caption={Proposition Generation Prompt},
                   label={lst:prop-gen}]
You are given descriptions of a desired reward function in English. Your job is to construct a set of propositions (high-level events) that will help decompose these tasks into sub-goals suitable for reward machine construction. When defining propositions, ensure that no two propositions are logical negations of each other.

Guidelines:
- Each proposition should describe an event or condition that the tasks refer to directly: the completion of an action a task names, or a state a task asks to reach or to avoid. Do not introduce propositions for preconditions or intermediate configurations needed to achieve them.
- Distinguish exactly the objects and attributes that the tasks distinguish, so that every sub-goal and constraint in the tasks can be written as a boolean formula over the propositions.
- Do not use aggregate or counting propositions; the reward machine keeps track of progress over time.

Please only include propositions that are necessary for representing the tasks.

Scene description:
{SCENE_DESCRIPTION}

These propositions should help represent the following tasks:
 - {TASK_1}
 - {TASK_2}
 ...

For each proposition, provide a short lowercase symbol (e.g. ab, bc) used in logical formulas and a brief description of what the proposition means.

Return ONLY the propositions, one per line, in this exact format:
<symbol> = <description>

Do not include any other text, headers, or explanations.
\end{lstlisting}

\subsection{Reward machine generation}

The decomposed prompt is shown below. \texttt{\{SCENE\_DESCRIPTION\}} is the environment description followed by its physical constraints (\Cref{lst:env-constraints}), and \texttt{\{PROPOSITIONS\}} is the reference or the generated proposition set.

\begin{lstlisting}[style=promptstyle,
                   caption={Reward Machine Generation Prompt (decomposed)},
                   label={lst:rm-gen}]
You are an expert reward machine designer.
Scene description:
{SCENE_DESCRIPTION}

You are given the following task:
 - {TASK}
And the set of propositions you can use to express the task with.
{PROPOSITIONS}

The presented task can be expressed using a set of reward machines. If the task involves sub-goals that can be achieved independently, express each as a separate reward machine. The sub-task reward machines will later be composed via cross-product. Two sub-goals are independent when neither has to be achieved before the other and neither's propositions constrain the other. Keep independent sub-goals in separate reward machines, even when each sub-goal has its own internal sequence of stages. Only combine behaviours into one reward machine when one stage must follow another within the same line of progress, or when a constraint must hold throughout the task. If a constraint must hold at every step throughout the task (e.g., 'never pick up X'), encode it as reject transitions within the main task RM rather than as a separate constraint RM. If two consecutive transitions carry the identical guard formula, collapse them into one: A->B with guard G followed by B->C with the same G becomes A->C with guard G. {SHARED_INSTRUCTIONS} Please return these reward machines.

Reward Machine states should be numbered uA, uR, u0, u1, u2, u3, ..., with uA always being the accepting state, uR always the rejecting state and u0 always the initial state. Transitions should be represented as a tuple (i, j, phi, r), where i and j are the start and end Reward Machine states of the transition, respectively, phi is a logical formula over the set of propositional symbols (use "~" to represent "not", "&" to represent "and", and "|" to represent "or", and write the formula in disjunctive normal form), and r is the reward for the transition. Your output should be the transitions in the Reward Machine, one per line, in the tuple form shown above, e.g. (u0, u1, ~X & Y, 0), with no other punctuation. For brevity, do not list self-loop transitions that provide 0 reward in the output. Rewards: use 0 for intermediate transitions, 1 for reaching uA, and -1 for any backward transition that leaves uA.

To parse this output easily, please return in the following format:
RM#1:
<transitions_go_here>
RM#2:
<transitions_go_here>
...

Do not include any reasoning, explanation, or commentary -- only the RM blocks.
\end{lstlisting}

\texttt{\{SHARED\_INSTRUCTIONS\}} is the following text, used by both the decomposed and the monolithic prompt:

\begin{lstlisting}[style=promptstyle,
                   caption={Shared RM design instructions},
                   label={lst:rm-shared}]
Assume the initial state requires the maximum number of steps to complete the task -- no sub-goal is already satisfied. If a precondition is already satisfied at runtime, the RM will simply skip that transition immediately. Decompose each task into the maximum number of meaningful intermediate states that track progress. Do not add shortcut transitions that skip intermediate states. When the task has a necessary logical precondition, please include it in the reward machine even if it is not directly described by the task description. When the environment states that one condition must hold before another can be achieved, give that precondition its own transition, reached before the transition it enables. Guard each backward transition with only the negation of the condition that established the state it leaves. If the environment describes a condition as a lasting state that can later become false again, include backward transitions from every state where that condition was established -- not just from the accepting state. Do not add backward transitions for conditions the environment describes as momentary or as one-off events. When multiple propositions together identify a single entity, use their conjunction in guards to distinguish individual entities. A guard mentioning only one property creates non-determinism if the omitted property matters for tracking progress. Transition guards from the same state must be mutually exclusive -- no valuation should satisfy more than one guard. For repeating or cyclic tasks, do not use an accepting state (uA). Instead, cycle back to the initial state and give +1 reward for completing each cycle. The cyclic RM should contain only the transitions needed for one cycle; do not add reject states unless the task description explicitly mentions something to avoid.
\end{lstlisting}

The monolithic prompt is identical except that the paragraph describing the decomposition is replaced by the following, and it asks for a single \texttt{RM\#1} block:

\begin{lstlisting}[style=promptstyle,
                   caption={Reward Machine Generation Prompt (monolithic): differing paragraph},
                   label={lst:rm-gen-mono}]
Express this task as a single reward machine. If the task involves sub-goals that can be achieved independently, the RM must handle all valid orderings of those sub-goals internally. Two sub-goals are independent when neither has to be achieved before the other and neither's propositions constrain the other. If a constraint must hold at every step throughout the task (e.g., 'never pick up X'), encode it as reject transitions within the RM. If two consecutive transitions carry the identical guard formula, collapse them into one: A->B with guard G followed by B->C with the same G becomes A->C with guard G. {SHARED_INSTRUCTIONS} Please return this reward machine.
\end{lstlisting}

\begin{lstlisting}[style=promptstyle,
                   caption={Physical constraints appended to the scene description},
                   label={lst:env-constraints}]
OGBench-Scene:
Physical constraints:
- The drawer/window must be unlocked before it can be opened or closed.
- The lock only freezes the handle, so it can be engaged or released whether the drawer/window is open or closed. Locking it while open leaves it open until it is unlocked again.
- The cube can only be put inside the drawer while the drawer is open, but once it is inside it stays inside even if the drawer is later closed.
- Locked/unlocked, open/closed and the cube's location are lasting states, and each can change back later: an opened drawer can be closed again, and a released lock can be re-engaged.

GeoGrid:
Physical constraints:
- The agent occupies a single cell, so it is on at most one object at a time: no two colours, and no two shapes, can hold in the same step.
- Every object has exactly one colour and exactly one shape, so a colour never holds without a shape and a shape never holds without a colour.
- Being on an object is momentary: it holds only while the agent stands on that object, and stops as soon as the agent moves to another cell.
- The valid observations are: nothing (empty cell), or exactly one colour-shape pair from {r&c, r&t, g&c, g&t, b&c, b&t}. A guard using only a colour (e.g., 'r') matches two distinct objects; use the conjunction (e.g., 'r&c' or 'r&t') when those objects must be tracked separately.
- Since observations are momentary, the last required visit should directly trigger the transition to uA. When multiple paths lead to uA, connect each path directly to uA with its own guarded transition rather than merging paths through an intermediate state.

MiniWorld-Pickup:
Physical constraints:
- The agent picks up at most one object at a time, so no two colours can hold in the same step, and a box and a ball cannot both hold in the same step.
- Every object has exactly one colour and is either a box or a ball, so a colour never holds without a box or ball and a box or ball never holds without a colour.
- There is exactly one box and one ball of each colour (6 objects total). After picking up one colour-kind pair (e.g., r&x), the only remaining object of that colour is the other kind (e.g., r&o), so the same conjunction cannot occur twice.
- Picking up an object is an event that happens in a single step, not a lasting state, and it cannot be undone: the object is removed from the room.
- The valid observations are: nothing (no pickup), or exactly one colour-kind pair from {r&x, r&o, g&x, g&o, b&x, b&o}. A guard using only a colour (e.g., 'r') matches two distinct objects; use the conjunction (e.g., 'r&x' or 'r&o') when those objects must be tracked separately.
- Since each pickup is instantaneous, a task is complete the moment the last required pickup occurs -- the pickup event itself should trigger the transition to uA.
- Since pickups are irreversible, when the task specifies a strict ordering, add reject transitions at each state for every pickup that is not the next expected one.
\end{lstlisting}

\subsection{Label prior generation}

The prompt receives the same scene description and physical constraints as RM generation (\Cref{lst:env-constraints}). The JSON template at the end is generated from the states and propositions of the given RMs.

\begin{lstlisting}[style=promptstyle,
                   caption={Label Prior Generation Prompt},
                   label={lst:prior-gen}]
You are an expert in reward machines and in tracking which reward machine (RM) state an agent is in from noisy observations.

{SCENE_DESCRIPTION}

Task:
 - {TASK}

Propositions:
{PROPOSITIONS}

The task has been decomposed into the following reward machines, which run in parallel, one per sub-goal. Each transition is (from_state, to_state, formula, reward); u0 is the initial state, uA the accepting state, and self-loops with 0 reward are omitted.

{REWARD_MACHINES}

Each RM tracks these propositions:
{RM_PROPOSITIONS}

We need to know, for every RM state u and every proposition p used by that RM, what holds at the time steps when that RM is in state u.
First classify every proposition as LASTING or MOMENTARY:
- LASTING: it describes a condition of the world that, once it holds, keeps holding until something changes it.
- MOMENTARY: it describes an event that holds only at the single time step at which the event happens.
A MOMENTARY proposition holds only at the step at which its event happens, which is the step that triggers the RM transition, so in every RM state u its value is FALSE and its next_step_change is YES. Answer the questions below in full for the LASTING propositions.

Then give, for every proposition, its value (TRUE or FALSE) at the start of the episode, in the situation the task describes (where the task says nothing, assume no sub-goal is already satisfied). Use it for the initial state u0 and for every state the RM can be in before that proposition has changed.

Answer two separate questions for each pair of an RM state u and a proposition p:
- value: Is p true at the moment the RM reaches u, either at the start of the episode or through a transition into u? Answer TRUE or FALSE.
- next_step_change: While the RM is in u, can a single action of the agent change the value of p at the next time step? Answer YES if this is possible in at least one situation the RM can be in while in u, even if the change would move the RM to another state. Answer NO if, in every such situation, changing p would first require another proposition to change in a way that moves the RM out of u, or if the physical constraints forbid changing p at all.

Reason from the physical constraints and from how the RM can reach u: consider every path into u, including the initial state and backward transitions.

In addition, at each time step the labelling function returns the set of propositions that are true. For each RM, list every combination of that RM's MOMENTARY propositions that can occur at a single time step, and no others. Write each combination as its true propositions joined by '&' (e.g. a&b), and the combination in which none of them is true as "empty". LASTING propositions are not part of this list (their values in each state are given by your answers above); write ["empty"] for an RM without MOMENTARY propositions. Check each combination against the physical constraints: a combination is possible unless a constraint rules it out.

Output a single JSON object inside a ```json code block with exactly this structure, replacing each "..." type with LASTING or MOMENTARY, each "..." value with TRUE or FALSE, each "..." next_step_change with YES or NO, each "..." reason with one short sentence, and each label_sets list with the label sets:
{JSON_TEMPLATE}
Output nothing after the code block.
\end{lstlisting}

\subsection{Proposition labelling}

The labelling prompt takes a numbered list of propositions, each given by its key and its question. In our experiments each query contains a single proposition. Each proposition comes with one reference frame where it is true and one where it is false.

\begin{lstlisting}[style=promptstyle,
                   caption={Proposition Labelling Prompt},
                   label={lst:prop-label}]
You are an expert at analyzing robot manipulation images.
{SCENE_DESCRIPTION}

For each frame, determine whether each of the following propositions is true (1) or false (0):
{PROPOSITION_LIST}

For every proposition, first write a brief analysis of the visual evidence before providing the binary value.

Output a single JSON object whose keys are frame indices (as strings) and whose values are objects mapping proposition keys to 0 or 1.
Example format: {"0": {"<prop>": 0 or 1}, "1": {"<prop>": 0 or 1}, ...}

Before the episode frames, here are reference examples for each proposition -- one frame where the proposition is TRUE and one where it is FALSE. Use these to anchor what the positive and negative states look like in this environment; the actual per-frame judgements below may still be either value.
Reference: [<prop>] = 1 (TRUE)
[IMAGE]
Reference: [<prop>] = 0 (FALSE)
[IMAGE]

Now the episode frames:
You are given {N} frames from a robot manipulation episode in temporal order.
Frame 0:
[IMAGE]
Frame 1:
[IMAGE]
...
\end{lstlisting}

\subsection{Progress estimation (GVL)}
\label{app:gvl-progress}

The following prompts are used for VLM-based progress estimation.
The GVL baseline asks for whole-task progress, and \themethoddense{} asks for progress toward each proposition (\Cref{tab:main_results,sec:reward-signal-ablation}).
For OGBench-Scene, \themethoddense{} also queries GVL for the progress toward each negated proposition, using a frame where the proposition holds as the 0\% reference.
In GeoGrid and MiniWorld-Pickup, $\vevt{\neg\proposition}(\statee) = 1 - \hat{\ell}_\proposition(\statee)$ is read from the grounding labels.
For MiniWorld-Pickup, \texttt{\{TASK\_DESCRIPTION\}} is the scene description (\Cref{lst:miniworld-scene}) followed by a \texttt{TASK:} line holding the task or proposition, and an image of the empty room is the 0\% reference frame.
Since a pickup requires first reaching the object, the per-proposition goal of \themethoddense{} is a matching object being close and centred, e.g.\ for \texttt{r} ``A red object (box or ball) is close and centered in front of the agent''.

\begin{lstlisting}[style=promptstyle,
                   caption={Progress Estimation Prompt (GVL)},
                   label={lst:progress-est}]
You are an expert roboticist tasked to predict task completion percentages for frames of a robot for the task of '{TASK_DESCRIPTION}'.
The task completion percentages are between 0 and 100, where 100 corresponds to full task completion.
Note that the following frames are in random order, so please pay attention to the individual frames when reasoning about task completion percentage.

[FRAMES]

Now, for the task of '{TASK_DESCRIPTION}', output the task completion percentage for the following frames that are presented in random order. For each frame, format your response as follows: Frame {i}:
Frame Description: {description}, Task Completion Percentage: {percentage}%

Do not omit any frame.
\end{lstlisting}

\subsection{Success detection (VLM-SD)}

The VLM-SD baseline asks for the first frame at which the whole task is completed. \texttt{\{TASK\_DESCRIPTION\}} is the scene description followed by the task's language command.
In MiniWorld-Pickup, the episode is shown at stride $8$.

\begin{lstlisting}[style=promptstyle,
                   caption={Success Detection Prompt (VLM-SD)},
                   label={lst:vlm-sd}]
You are an expert at analyzing robot manipulation trajectories.

{TASK_DESCRIPTION}

This task requires observing the sequence of frames to determine its completion.
Carefully reason whether the complete task has been completed at each frame.

You are given {N} frames from a robot manipulation episode in temporal order. For each frame, analyze whether ALL sub-goals of the task have been achieved.

Output a single JSON object:
  {"first_success_frame": <int>}
where the value is the 0-based frame index of the FIRST frame where the task is fully completed. If the task is never completed in the episode, return:
  {"first_success_frame": -1}
Frame 0:
[IMAGE]
Frame 1:
[IMAGE]
...
\end{lstlisting}

\section{Use of AI-Assisted Technologies}
\label{app:ai-use}

The authors used Claude models (Anthropic) through Claude Code as an interactive assistant for writing and refactoring experiment code, including figure-generation scripts, launching and monitoring runs, aggregating results, drafting and editing text, and writing the Matplotlib code used to render \Cref{fig:method-overview}.
All methods, experiments, claims and final text were designed, decided and verified by the authors, who take full responsibility for the content.

\end{document}